\documentclass[10pt,twocolumn,letterpaper]{article}

\usepackage{cvpr}              

\usepackage{graphicx}
\usepackage{amsmath}
\usepackage{amssymb}
\usepackage{booktabs}
\usepackage{algorithm}
\usepackage{algpseudocode}
\usepackage{multirow}
\usepackage{placeins}

\usepackage{color}
\usepackage[dvipsnames]{xcolor}

\usepackage{xspace}
\usepackage{float}
\usepackage{contour}
\usepackage{xkeyval}

\usepackage[pagebackref,breaklinks,colorlinks]{hyperref}
\usepackage[abs]{overpic}

\usepackage[capitalize]{cleveref}
\crefname{section}{Sec.}{Secs.}
\Crefname{section}{Section}{Sections}
\Crefname{table}{Table}{Tables}
\crefname{table}{Tab.}{Tabs.}
\newcommand{\ouralg}{FvOR\xspace}

\contourlength{.4pt}
 \newcommand{\figlabel}[1]{\sffamily\bfseries\footnotesize\contour{white}{#1}} 
\makeatletter
\newlength{\sfp@hseplen}\newlength{\sfp@vseplen}
\define@cmdkey{subfigpos}[sfp@]{vsep}[0.6\baselineskip]{\setlength{\sfp@vseplen}{\sfp@vsep}}
\define@cmdkey{subfigpos}[sfp@]{hsep}[40pt]{\setlength{\sfp@hseplen}{\sfp@hsep}}
\newcommand{\subfigimg}[3][,]{%
  \setkeys{Gin,subfigpos}{vsep,hsep,#1}
  \setbox1=\hbox{\includegraphics{#3}}
  \leavevmode\rlap{\usebox1}
  \rlap{\hspace*{\sfp@hsep}\raisebox{\dimexpr\ht1-80pt}{\figlabel{#2}}}
  \phantom{\usebox1}
}
\makeatother

\def\cvprPaperID{5429} 
\def\confName{CVPR}
\def\confYear{2022}

\let \bs=\mathbf
\let \set=\mathcal
\def \gt {\textup{gt}}
\def \image {\textup{image}}
\def \conv {\textup{3D}}

\begin{document}

\title{\ouralg: Robust Joint Shape and Pose Optimization for Few-view Object Reconstruction}
\author{Zhenpei Yang\\
The University of Texas at Austin \\
{\tt\small yzp@utexas.edu}
\and
Zhile Ren\\
Apple\\
{\tt\small zhile\_ren@apple.com} 
\and
Zaiwei Zhang\\
The University of Texas at Austin \\
{\tt\small zaiweizhang@utexas.edu}
\and
Miguel Angel Bautista Martin\\
Apple\\
{\tt\small mbautistamartin@apple.com}
\and
Qi Shan\\
Apple\\
{\tt\small qshan@apple.com}
\and
Qixing Huang\\
The University of Texas at Austin \\
{\tt\small huangqx@cs.utexas.edu}
}

\author{
  \hspace{-1.3cm}
  \begin{tabular}[t]{c}
    Zhenpei Yang$^{1}$ \quad  Zhile Ren$^2$ \quad Miguel Angel Bautista$^2$\\ \quad Zaiwei Zhang$^{1}$ \quad Qi Shan$^2$ \quad Qixing Huang$^{1}$\\
    $^1$The University of Texas at Austin \quad $^2$Apple\\
\end{tabular}
}
\maketitle
\let\thefootnote\relax\footnotetext{$^*$ Experiments are conducted by Z. Yang at the University of Texas at Austin. Email: yzp@utexas.edu}

\renewcommand{\baselinestretch}{0.96}

\begin{abstract}
Reconstructing an accurate 3D object model from a few image observations remains a challenging problem in computer vision. State-of-the-art approaches typically assume accurate camera poses as input, which could be difficult to obtain in realistic settings. In this paper, we present \ouralg, a learning-based object reconstruction method that predicts accurate 3D models given a few images with noisy input poses. The core of our approach is a fast and robust multi-view reconstruction algorithm to jointly refine 3D geometry and camera pose estimation using learnable neural network modules. We provide a thorough benchmark of state-of-the-art approaches for this problem on ShapeNet. Our approach achieves best-in-class results. It is also two orders of magnitude faster than the recent optimization-based approach IDR~\cite{idr}. Our code is released at \url{https://github.com/zhenpeiyang/FvOR/}.

\end{abstract}

\section{Introduction}
\label{sec:intro}

Reconstructing the 3D shape of objects solely from unregistered RGB inputs is a long-standing problem in computer vision. One popular pipeline is to integrate Structure-from-Motion (SfM) and Multi-view Stereo (MVS)~\cite{Ma:2003:3D,10.5555/861369}. A common principle of this popular pipeline is to recover relative camera poses, establish pixel correspondences (either explicitly or implicitly), and solve triangulation to obtain a dense reconstruction. The success of this paradigm relies on dense image coverage to obtain accurate camera poses and correspondences~\cite{10.1145/1141911.1141964,DBLP:journals/pami/FurukawaP10,DBLP:conf/cvpr/FurukawaCSS10,DBLP:journals/cacm/AgarwalFSSCSS11}. 
Enabled by the emergence of large scale 3D datasets that provide shape priors about 3D objects,
a recent line of works focus on learning monocular 3D reconstruction~\cite{DBLP:conf/nips/EigenPF14,DBLP:conf/iccv/EigenF15,DBLP:conf/cvpr/GodardAB17,DBLP:conf/eccv/0001XLTTTF16,choy20163d,DBLP:conf/nips/0001WXSFT17}.
The general idea is to learn multi-scale correlation priors among different regions of geometric shapes, which are used to infer complete geometry from partial observations.

\begingroup
\begin{figure}[!t]
 \def\imh{0.13\textheight}
 \def\imhs{0.05\textheight}
 \def\imw{0.13\textwidth}
 \def\imww{0.08\textwidth}
  \def\imwt{0.3\textwidth}
  \def\imht{0.16\textheight}
\begin{center}
 \setlength{\tabcolsep}{3pt}
\begin{tabular}{ccccc}
\includegraphics[width=\imww, keepaspectratio]{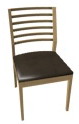} & \includegraphics[width=\imww, keepaspectratio]{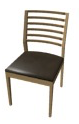} & \includegraphics[width=\imww, keepaspectratio]{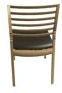} & \includegraphics[width=\imww, keepaspectratio]{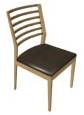} & \includegraphics[width=\imww, keepaspectratio]{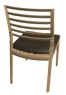} \\
\multicolumn{5}{c}{\includegraphics[width=\imw,keepaspectratio]{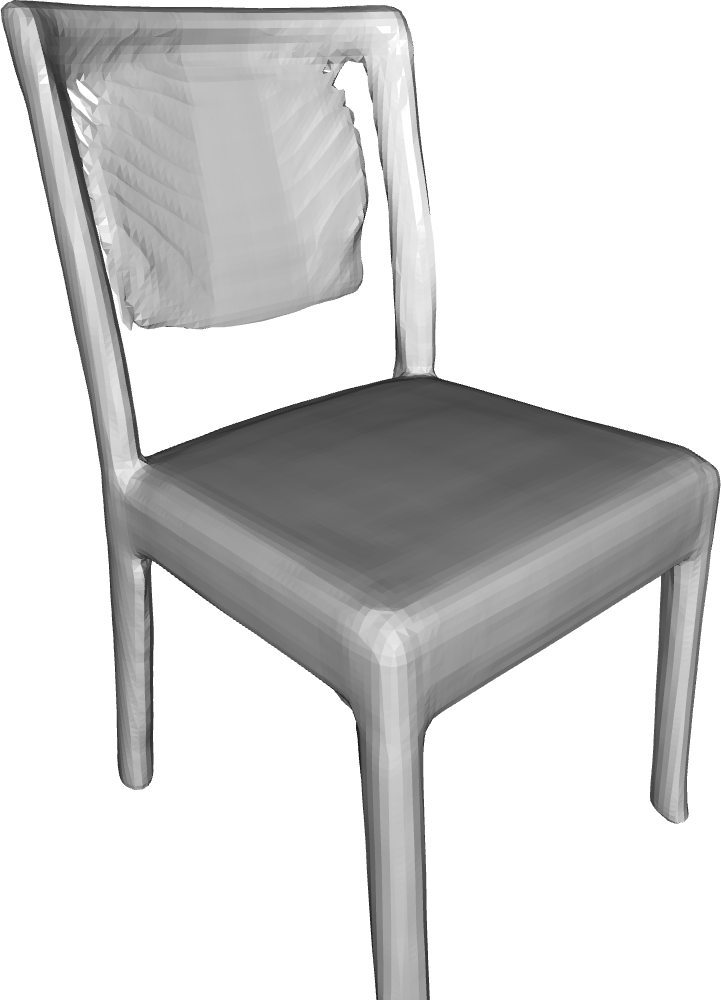} \quad \includegraphics[width=\imw,keepaspectratio]{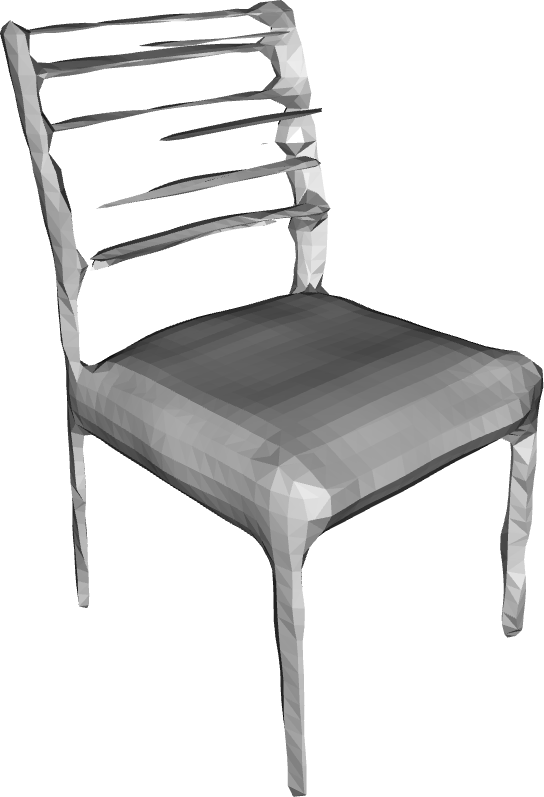} \quad \includegraphics[width=\imw,keepaspectratio]{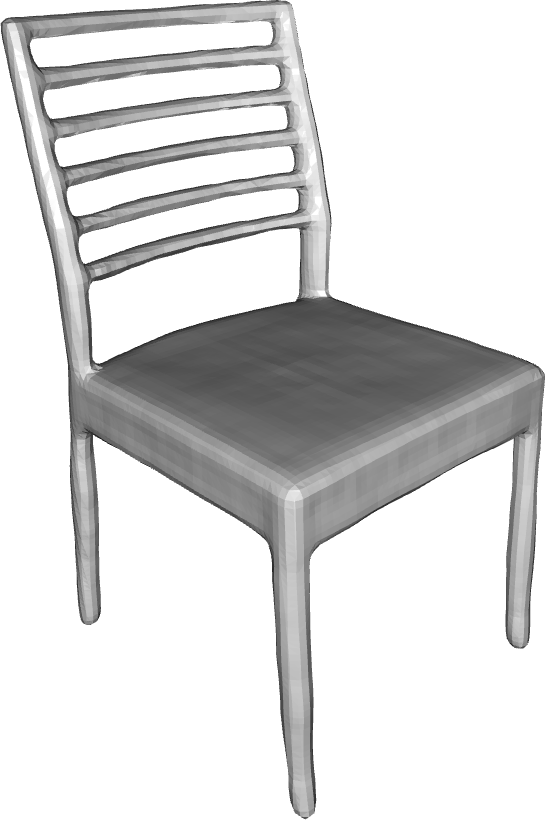}} \\

\multicolumn{5}{c}{OccNet$^\dagger$~\cite{mescheder2019occupancy,apple} \quad  \quad \quad  IDR~\cite{idr} \quad  \quad \quad \ouralg (Ours)}

\end{tabular}
\vspace{-0.0in}
\caption{Our approach \ouralg outperforms state-of-the-art approaches of few-view 3D reconstruction. } 
\end{center}
\vspace{-0.3in}
\end{figure}
\endgroup

\begin{figure*}[!t]
\centering
\vspace{0.1in}
\begin{overpic}[width=0.8\textwidth]{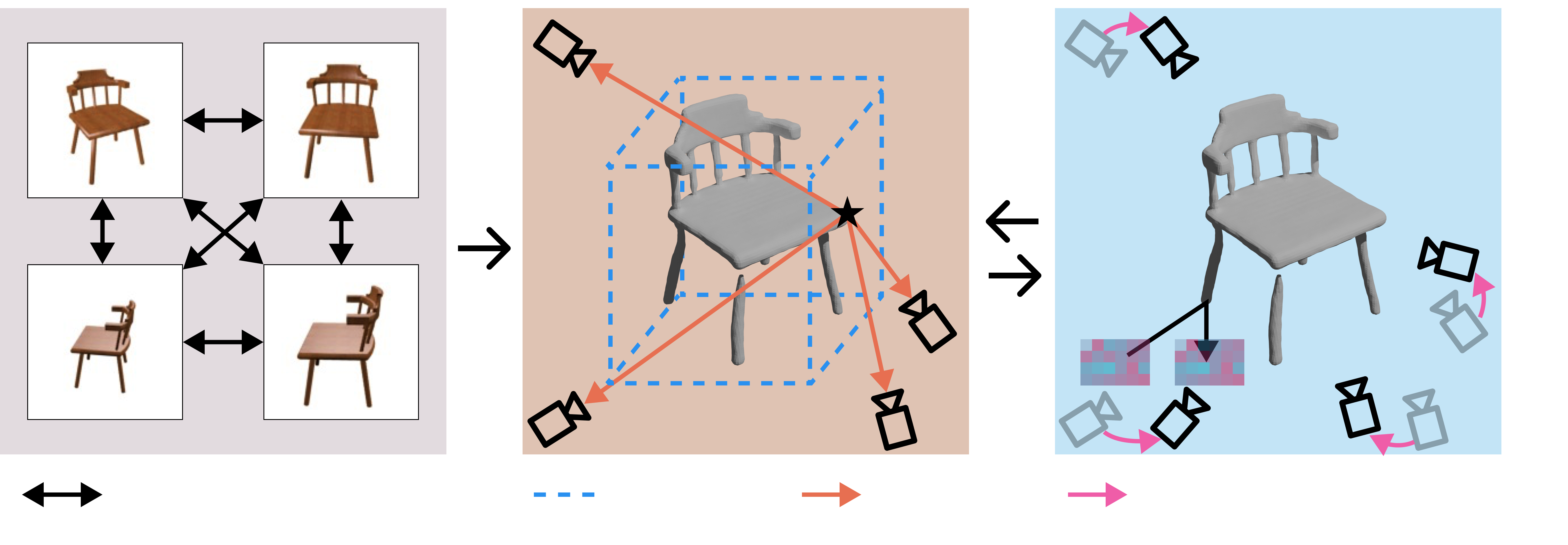}
\put(30,11){\footnotesize{Cross Image Attention}}

\put(38,140){Pose Init}

\put(160,140){Shape Update}

\put(297,140){Pose Update}

\put(154,11){$f_{\text{3D}}$}
\put(220,11){$f_{\text{image}}$}

\put(221,83){$\hat{g}(x)$}

\put(315,35){$\hat{f_i}$}
\put(265,40){$f_i$}

\put(290,11){\footnotesize{Levenberg–Marquardt step}}
\vspace{-0.2in}
\end{overpic}

\vspace{-0.2in}
\caption{Our approach consists of two stages. The first stage is \textit{pose initialization} which predicts an initial pose for each input image. The second stage alternates between \textit{shape update} and \textit{pose update} to give an accurate reconstruction with jointly improved camera poses. }
\vspace{-0.15in}
\label{fig:workflow}
\end{figure*}
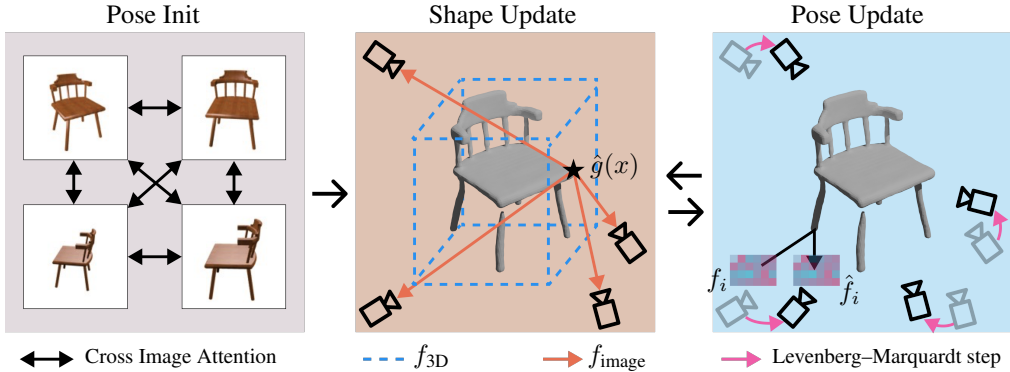

Acquiring dense input views is crucial for achieving good 3D reconstruction quality on current pipelines, but it is also a very tedious and not user-friendly process. For instance, a casual non-expert user that just began using 3D reconstruction applications (such as creating 3D models of their house), may overlook the strict requirements of capturing high-quality dense views.

In this paper, we study the setting of few-view reconstruction~\cite{choy20163d}, which sits between dense-view reconstruction and single-view reconstruction. The promise of this setting is that the input views cover the most of underlying object, and one only needs to fill in a small portion of missing regions, a task that is easier to achieve than single-view reconstruction. The ultimate goal is to match the quality of dense reconstruction while significantly reducing the number of inputs. While both few-view reconstruction and single-view reconstruction fall into the category of learning-based approaches, the performance of few-view reconstruction relies on accurate image poses, which could be challenging to estimate from the input images themselves in realistic scenarios. In dense-view reconstruction, the SfM pipeline estimates image poses by first predicting relative camera poses using feature correspondences and then performing synchronization~\cite{DBLP:conf/iccv/ChatterjeeG13,Crandall:2013:SFM} to extract absolute camera poses. However, this pipeline does not apply to few-view reconstruction as there are only a few images, which makes the accurate pose prediction using correspondence difficult.

This paper introduces a novel learning-based approach for joint optimization of the shape reconstruction and the camera poses associated with the input images. The core of our approach consists of a pose initialization module, a shape module, and a pose refinement module. The pose initialization module computes an initial camera pose for each input image. The shape and pose refinement modules are alternated to improve the shape reconstruction and the camera poses jointly. We design the pose initialization module using a geometric approach, aiming to reduce outlier predictions of camera poses that are difficult to rectify in pose refinement. The shape module combines the strengths of per-view image features and 3D convolutional features to obtain an accurate implicit 3D reconstruction with shape details. The pose refinement module performs geometric alignments between rendered images and real images in a learned feature space. Both the shape and pose modules are end-to-end trainable. Compared to existing learning-based image pose estimation techniques, our approach uses a dynamically changing 3D reconstruction and geometric constraints, both of which are unavailable in standard end-to-end pose estimation approaches~\cite{DBLP:conf/cvpr/TulsianiM15,DBLP:conf/rss/XiangSNF18,9543652,DBLP:journals/corr/abs-2105-14291}.

Our approach achieves state-of-the-art results on ShapeNet. The shape reconstruction module also improves upon state-of-the-art approaches under the setting of known camera poses. Due to the efficiency of our neural network modules, our approach is two orders of magnitude faster than the recent optimization-based approach IDR~\cite{idr}.

\section{Related Work}

\noindent\textbf{Single-view Object Reconstruction}  Typical single-view approaches use an image encoder to estimate a latent code, which is then decoded into 3D shape representations such as voxels~\cite{girdhar2016learning}, point-clouds~\cite{fan2017point}, meshes~\cite{wang2018pixel2mesh,groueix2018papier}, skeletons~\cite{kar2015category,wu2016single}, or implicit functions~\cite{mescheder2019occupancy}. Although this methodology has shown promising results, they are inherently limited to large uncertainties in the invisible regions given the partial visible observations (c.f.~\cite{what}). 

\noindent\textbf{Dense-view Object Reconstruction} Traditional approaches for reconstructing an object usually involve dense scanning around the object, followed by SfM(Structure from Motion)~\cite{Snavely2010,wu2011visualsfm,Crandall:2013:SFM,snavely2006photo} or SLAM(Simultaneous Localization and Mapping)~\cite{durrant2006simultaneous,yang2019cubeslam} approaches for reconstruction and camera pose estimation. Popular software includes COLMAP~\cite{schoenberger2016sfm} and OpenMVG~\cite{moulon2016openmvg}. Deep learning counterparts, for example DeepV2D\cite{teed2018deepv2d, bloesch2018codeslam}, have also been proposed in recent years. ~\cite{lin2019photometric} proposed an approach to recover object shape from posed video frames by combing a deep shape prior network with photometric optimization. Recently, the seminar work NeRF~\cite{mildenhall2020nerf} inspire many research learning implicit 3D representation from images.  Some recent works~\cite{idr,lin2021barf,meng2021gnerf} consider inaccurate camera poses as input and optimize the shape reconstruction~\cite{idr}(or neural radiance field~\cite{lin2021barf,meng2021gnerf}) and poses jointly. This type of approach requires the most expensive capture efforts, but usually gives good performance. 

\noindent\textbf{Few-view Object Reconstruction} 
Such a task aims at reconstructing the underlying object given several images. In general, there are two approaches to solving this task: whether they model camera poses explicitly during the inference time, or not. One pioneering work of the pose-free approach is 3D-R2N2~\cite{choy20163d}, which uses a 3D convolutional LSTM to aggregate multi-view information sequentially. A recent example of this type of approach is Pix2Vox++~\cite{xie2020pix2vox++}. The other approaches model camera poses directly. Many of these approach assume ground truth camera poses as input~\cite{lmvs,apple,peng2020neural,oechsle2021unisurf,wang2021neus,yang2021deep,yu2021pixelnerf,disn}. For example, ~\cite{lmvs} use ground truth camera pose to build a volumetric feature representation, which is then decoded into discrete voxel. \cite{yang2021deep} proposed learning a shape prior during training, and optimizing the shape code to minimize silhouette loss during testing to recover the shape. ~\cite{disn} directly uses predicted camera pose obtained from pre-trained network. Recently, NeRS~\cite{zhang2021ners} proposed a NeRF-style few-view reconstruction method using a neural surface representation. Our approach innovates in learning shape reconstruction and pose estimation using deep learning. As a result, our approach does not require object masks~\cite{idr,zhang2021ners} or category-specific mesh initialization~\cite{zhang2021ners}. Moreover, our approach requires only a few updates, and the running time is significantly faster than IDR~\cite{idr} and NeRS~\cite{zhang2021ners}.

\section{Approach}

We first introduce the problem statement and an overview of our approach in Sect.~\ref{sec:problem}. We then elaborate on the technical contributions from Sect.~\ref{sec:pose-init} to Sect.~\ref{Section:Alternate}.
\subsection{Few-view Object Reconstruction}
\label{sec:problem}

\noindent\textbf{Problem statement.} Given a set of RGB images  $\mathcal{I} = \{I_i\ | i=0,\dots, k - 1\}$ observing a single object, where $k$ is the number of observations, we aim to recover the 3D mesh model $S$ of the underlying object, up to a global similarity transformation. We assume the camera intrinsic matrix $K$ is known and fixed across all views. 

\noindent\textbf{Approach overview.} Fig.~\ref{fig:workflow} is an overview of \ouralg. It starts with a pose initialization module that predicts camera poses for each image. This module gives us initial pose estimates with acceptable accuracy. We then alternate between reconstructing the shape from input images with current poses and performing image-shape alignment to refine the poses of each input image. For the shape reconstruction module, we combine a two-stream network that integrates image-based features with 3D features. Image-shape alignment is performed in a learned feature space between input images and corresponding rendered images of the predicted shape. Both modules are end-to-end differentiable. We alternate between the shape and pose modules to reconstruct an accurate 3D model from few-view inputs. 

A common approach for training the alternating mechanism is to stitch the alternating shape, pose modules together, and enforce a loss on the final output. We found that this strategy is challenging to train and is not very flexible. In the same spirit as the gradient operators in alternating minimization, this paper trains each module in isolation while forcing them to make progress under different inputs. For example, the pose module is learned to recover the underlying ground truth under randomly perturbed poses. This methodology offers excellent flexibility in developing training losses and instilling training data.  

\subsection{Pose Initialization Module}\label{sec:pose-init}

The goal of the pose initialization module is to provide initial camera poses for subsequent shape and pose optimization steps. As the camera poses can be refined later, we design a pose initialization module to reduce the number of pose outliers, which are hard to rectify later in the pose refinement stage. For each pixel of each input image, we predict its 3D coordinate in a world coordinate system (scene coordinate) of the underlying geometry~\cite{shotton2013scene, nocs2019}. The pose is then obtained by performing global matching between 2D image pixels and the corresponding 3D points via RANSAC~\cite{journals/cacm/FischlerB81}. Our approach exhibits three advantages compared to existing regressing and classification based pose estimation approaches~\cite{9543652,DBLP:journals/corr/abs-2105-14291}. First, reconstructing the 3D coordinates of each image uses information from all input images during testing, meaning the camera poses are jointly predicted. Second, pose regression enforces geometric constraints between correspondences. Third, RANSAC can efficiently deal with incorrect 3D coordinates. 

\noindent\textbf{Scene coordinate prediction.} 
Our model first encodes a 2D feature map for each input image independently. We then use a multi-image attention module to aggregate features from all input images. Inspired by \cite{sun2021loftr,katharopoulos2020transformers}, the multi-image attention module is composed by alternating between self-attention and cross-attention blocks. The final output is a 3D coordinate $\hat{p}_{i,j}$ for each pixel. The detailed network design can be found in the supp. Network training minimizes the $l_2$ distances between the predicted and ground-truth scene coordinates. The loss for one set of input images is given by
$$
\mathcal{L}_{\text{init}} =  \sum_{i=0}^{k-1}\sum_{j=0}^{h\times w-1}w_{i,j}\big\|\hat{p}_{i,j} - d_{i,j}^{\gt} (R_i^{\gt}K^{-1} \small{\begin{pmatrix}
u_{i,j} \\
v_{i,j} \\
1 \\
\end{pmatrix}}+t_i^{\gt}) \big\|_2
$$
where $(u_{i,j},v_{i,j})$ is the pixel coordinate of the $j$-th pixel of the $i$-th input image; $d_{i,j}^{\gt}$ is its ground-truth depth; $T_i^{\gt}:=(R_i^{\gt}|t_i^{gt})\in SE(3) $ is the ground-truth camera to world pose of the $i$-th image. $w_{i,j}$ is a binary weight indicating whether or not this pixel has G.T. depth.

\noindent\textbf{Pose regression.} After we acquired scene coordinate estimates, we use an off-the-shelf RANSAC PnP approach to recover the pose estimates for each input view (details is in the supp.). This initial camera pose serves as input for the subsequent shape optimization module.

\subsection{Shape Optimization Module}\label{sec:joint-shape}

The shape optimization module takes as input the input images and their pose estimates $\{(I_i, \hat{T}_i^t)| i = 0,\dots, k -1\}$ and outputs a shape reconstruction. Motivated by the success of implicit shape representations~\cite{DBLP:conf/cvpr/ParkFSNL19,DBLP:conf/cvpr/ChenZ19}, we encode the 3D reconstruction as a deep signed distance function $\hat{g}^t: \mathcal{R}^3 \rightarrow \mathcal{R}$~\cite{DBLP:conf/cvpr/ParkFSNL19} that outputs the signed distance of any query point in the space. 

Our approach innovates computing the implicit function value $\hat{g}^t(\bs{x})$ by fusing features from two sources:
$$
\hat{g}^t(\bs{x}) = g_{\Theta}\big(\bs{f}_{\image}^t(\bs{x}), \bs{f}_{\conv}^t(\bs{x})\big),
$$
where $g_{\Theta}$ is a multi-layer fully connected network. Similar to \cite{disn,apple,yu2021pixelnerf,saito2019pifu,henzler2021unsupervised,raj2021pixel,kwon2021neural}, $\bs{f}_{\image}^t(\bs{x})$ is given by the features extracted from projecting $\bs{x}$ onto the input images: 
$$
\nonumber \bs{f}_{\image}^t(\bs{x}) = \text{Pooling}\big(F_i({\mathcal{P}_i(\bs{x}, T_i)})\big),
$$
where ${\mathcal{P}_i(\bs{x}, T_i^t)}$ is the projection of $\bs{x}$ on image $I_i$ given the current camera pose $T_i^t$; $F_i$ is augmented ResNet18~\cite{DBLP:conf/cvpr/HeZRS16} that takes each image as input and outputs the pixel-wise feature map;
\text{Pooling} is the average pooling function. In addition, $\bs{f}_{\conv}^t(\bs{x})$
 represent features obtained from a 3D feature volume:
\begin{equation}
\nonumber \bs{f}_{\conv}^t(\bs{x}) = V(\bs{x}|\bs{f}^t_{\image}, \Phi),
\end{equation}
where $V\in \mathcal{R}^{c\times d\times d \times d}$ is a 3D volume produced by a convolutional 3D U-Net~\cite{cciccek20163d} with trainable parameters $\Phi$. The input to this  3D U-Net is an initial volume built by evaluating $\bs{f}_{\image}^t(\bs{x})$ where $\bs{x}$ is the coordinates at $d\times d\times d$ grid position (See the supp.). 

\noindent\textbf{Network training.}
In addition to supervising the implicit shape reconstruction using ground-truth signed distance values, we also force the gradient field of the implicit representation to match the corresponding ground truth:
$$
\small
\min\limits_{g}\sum_{x_i\in \mathcal{S}_0}\big\|g(x_i) - s_i^{\text{gt}}\big\|_1\\
    + \lambda_{\text{grad}}\sum_{x_i\in \mathcal{S}_1}\Big\|\frac{\partial g}{\partial x_i}/\|\frac{\partial g}{\partial x_i}\| - n_i^{\text{gt}}\Big\|_2
$$
where $\set{S}_0$ are points sampled in the 3D space as done in DeepSDF~\cite{DBLP:conf/cvpr/ParkFSNL19}; $\set{S}_1$ are points in on the surface of the underlying object.

\subsection{Pose Optimization Module}\label{sec:joint-pose}

We now describe the module for updating the camera pose estimates given the current 3D reconstruction. Specifically, the input consists of the input images $\mathcal{I}$, the current implicit shape representation $\hat{g}^t$, and the current camera poses $\{\hat{T}_i^t | i = 0,\dots, k - 1\}$. The output of this module is the pose updates $\Delta \hat{T}^{t}=\{\Delta \hat{T}_{i}^{t}\}_{0\leq i < k}$ of the corresponding camera poses. Our key idea is to perform geometric alignment between the 3D reconstruction and the input images on a learned feature representation. This is achieved by rendering the 3D reconstruction and then aligning features extracted from the rendered images and the corresponding input images. The training objective enforces that the pose updates derived from these modules match the underlying ground truth. We now describe the technical details. 

\noindent\textbf{Efficient renderer.} 
The efficiency of end-to-end learning of the pose module depends on the efficiency of the implicit function renderer. Therefore, unlike IDR~\cite{idr} that repeatedly evaluates the implicit function to find the accurate intersection point, we use a volumetric grid of size $d\times d\times d$ ($d=64$ in our implementation) to discretize the implicit function and then render the discretized implicit function. Rendering is achieved using sphere tracing ~\cite{hart1996sphere,jiang2020sdfdiff}. 
Such discretization allows us to render $224\times 224$ images at $79.1$ FPS (IDR's speed is only $0.32$) using a Nvidia V100 GPU. The output of this module is a depth map which is converted into a 2D object mask $\hat{M}^{t}$ and a set of 3D points that represent the visible region of the current 3D reconstruction.

\noindent\textbf{Learning feature alignment.}
The goal of this sub-module is to find an incremental pose update $\Delta T_i^{t} \in SE(3)$ to better align the rendered object mask $\hat{M}_i$ and the input image $I_i$. The goal is to ensure that this sub-module is end-to-end trainable while utilizing as much information as possible. Instead of directly aligning $\hat{M}_i$ and $I_i$, we use a neural network to compute a dense feature space to align $\hat{M}_i$ and $I_i$. Since the underlying pose between $\hat{M}_i$ and $I_i$ is expected to be small, we found that it is sufficient to enforce the loss on pose update derived from aligning the corresponding points in the feature space. 

Specifically, let $f_{\Theta}$ denote the network that computes the dense image descriptor, and $\hat{f}_i = f_{\Theta}( \hat{M}_i)$ and $f_i = f_{\Theta}( I_i)$ be the resulting feature map. Denote current camera pose for image $i$ as $T_i^{0}:=(R_i^{0}|t_i^{0})$, and its corresponding G.T. as $T_i^{\gt}:=(R_i^{\gt}|t_i^{\gt})$. We employ the exponential map parametrization of the pose correction $\Delta T_i^{t} =\begin{pmatrix} 
e^{[\Delta c_i]_{\times}} & \Delta t_i\\ 
  0 & 1
\end{pmatrix}$. Let $\set{P}_i = \{p_j\}$ collect the 3D points of $I_i$ derived from rendering.  We propose to compute $\bs{c}_i$ and $\bs{t}_i$ by solving following non-linear least square problem: 
\begin{align}
    \min_{\Delta c_i, \Delta t_i} &  \sum_{p_j\in 
\set{P}_i}\|f_i(P(K\bs{p}_{j'})) - \hat{f}_i(u_j, v_j)\|_{\mathcal{F}}^2, \\
   \nonumber   \begin{pmatrix} \bs{p}_j'\\ 
  1
\end{pmatrix} & =  \begin{pmatrix} I+[\Delta c_i]_{\times} & \Delta t_i\\ 
  0 & 1
\end{pmatrix}  \begin{pmatrix} \bs{p}_j\\ 
  1
\end{pmatrix}, \\ 
\nonumber \bs{p}_j & = d_j K^{-1} (u_j, v_j, 1)^{T}.
\end{align}
Here $(u_j,v_j)$ denotes the pixel coordinates of $p_j$ in the rendered image. $P(K\bs{p}_{j'})$ denotes the corresponding pixel coordinates on the input image after applying the pose update. 

To obtain an explicit expression of $\Delta c_i$ and $\Delta t_i$,
we use a linear approximation. This leads to the following expression, which applies one-step of Levenberg–Marquardt~\cite{GVK502988711}: 
\begin{equation}
    (\Delta c_i, \Delta t_i)^{T} = -(J^{T}J + \lambda I)^{-1} ( J^{T} r),
\end{equation}
where $r  = \sum_{j} r_{j}$;$J = \sum_{j} J_{j}$;$I$ is the identity matrix;$\lambda$ is a constant. Moreover, $r_j$ and $J_j$ are given by
\begin{align}
    r_{j} &= f_i(P(Kp_j')) - \hat{f}_i(u_j, v_j)\\
    \nonumber J_j &= \frac{\partial r_j}{ \partial{(\Delta c_i, \Delta t_i)^{T}}} \\
    \nonumber &= \frac{\partial r_j}{\partial f_i}
    \frac{\partial f_i}{\partial{(u'_i,v'_i)}}
    \frac{\partial{(u'_i,v'_i)}}{\partial{P}}
    \frac{\partial{P}}{\partial p'_j} 
    K
    \frac{\partial p'_j}{(\Delta c_i, \Delta t_i)^{T}}
\end{align}

\begin{algorithm}[!t]
\caption{\ouralg 3D Recon Algorithm}\label{alg}
\begin{algorithmic}
\State{Input: $\mathcal{I}$, $\hat{\mathcal{T}}_0$}
\State{$\hat{g}^0 \gets \text{shape\_update}(\mathcal{I},\hat{\mathcal{T}}^0)$}
\For{$t=0:n_1$}
        \For{$j=0:n_2$}
        \State{$\hat{M}^t \gets $ render($\hat{g}^{t}, \hat{\mathcal{T}}^{t}$)}
        \State{$\hat{\mathcal{T}}^t \gets \text{pose\_update}(\mathcal{I}, \hat{g}^{t}, \hat{\mathcal{T}}^t, \hat{M}^t)$}
    \EndFor
    \State{$\hat{g}^{t+1} \gets \text{shape\_update}(\mathcal{I},\hat{\mathcal{T}}^{t})$}
    \State{
    $\hat{\mathcal{T}}^{t+1}
    \gets 
    \hat{\mathcal{T}}^t$
    }
\EndFor
\State{ Return $\hat{g}^{n_1}, \hat{\mathcal{T}}^{n_1}$}
\end{algorithmic}
\end{algorithm}

\noindent\textbf{Training Details} For training the pose refinement module, we simulate input poses by adding random perturbations to the ground truth camera poses~\cite{lin2021barf}. We then train the pose refinement module by forcing it to recover from the perturbations. We then minimize the following loss:
\begingroup
$$
\setlength\arraycolsep{2pt}
    \mathcal{L}_{\text{pose\_refine}} = \sum_{i}\Big\| \begin{pmatrix} I + [\Delta c_i]_{\times} & \Delta t_i\\ 
  0 & 1
\end{pmatrix} \cdot \begin{pmatrix} R_i^0 & t_i^0\\ 
  0 & 1
\end{pmatrix} -  \begin{pmatrix} R_i^{\gt} & t_i^{\gt}\\ 
  0 & 1
\end{pmatrix}\Big\|_{\mathcal{F}}^2
$$
\endgroup

\subsection{Alternating Shape and Pose Optimization}
\label{Section:Alternate}

We first train the shape module~\ref{sec:joint-shape} using GT poses. Then, we add noise to the GT poses to fine-tune the shape module and use the shape module's prediction on the fly to train the pose update module \ref{sec:joint-pose} with a single step update. At inference time, we make two modifications: first, we alternate between shape update and pose update for multiple iterations instead of the single iteration used in training. Second, we add a regularization term that penalizes very large deviations from initial pose estimates. The complete algorithm is shown in Algorithm \ref{alg}. We set $n_1=3$ and $n_2=5$ in our experiments.

\section{Experimental Results}
\label{Section:Results}

In this section, we present our experimental results. We first describe the datasets used for evaluation in Sect.~\ref{sec:dataset}. Then, we introduce the baselines for camera pose estimation (Sect.~\ref{sec:baseline-pose}) and 3D reconstruction (Sect. ~\ref{sec:baseline-shape}). In Sect.~\ref{sec:metric}  we discuss the evaluation metrics. Finally, we provide an analysis of results in Sect. \ref{sec:analysis}

\subsection{Datasets}\label{sec:dataset}

\noindent\textbf{ShapeNet.} This dataset was introduced by 3D-R2N2~\cite{Choi2015} based on ShapeNet~\cite{shapenet} and has become a widely used benchmark for single/multi-view 3D reconstruction. It contains objects from 13 categories from ShapeNet v1 \cite{shapenet}. For each object, it contains 24 views from a camera pointing to the origin, and has large azimuth variation but small elevation variation. We follow the training/test splits and evaluation protocol in~\cite{apple} and randomly sample 5 views out of the 24 views to form an input set.

\subsection{Baselines for Pose Initialization}\label{sec:baseline-pose}
In this section, we describe different baseline approaches for initializing the pose estimates.

\begin{table}
\centering
\footnotesize
\resizebox{0.45\textwidth}{!}{%
\begin{tabular}{rcccc}
\toprule
& DISN~\cite{disn} & Cai~\etal~\cite{cai2021extreme}& \ouralg-Quat & \ouralg \\ \hline

Base & 3.66 & 5.10& 4.46 & 3.82 \\ 
Base + Cross & \textbf{2.46} & \textbf{2.25}& \textbf{3.06} & \textbf{1.40}\\ 
\bottomrule
\end{tabular}
}
\vspace{-0.1in}
\caption{Ablation study of the pose initialization module on ShapeNet. The results are the Pixel Error$\downarrow$. We can see that cross-attention can help predict more accurate image poses. ``Base" means per-image prediction is used. ``Base+Cross" means that each image's feature map  interacts with other images through cross-image attention. }
\label{table:ablation-pose-init}
\vspace{-0.1in}
\end{table}

\begin{table}
\centering
\footnotesize
\resizebox{0.45\textwidth}{!}{%
\begin{tabular}{ccccc}
\toprule
     Metrics & All & w/o $f_{\text{image}}$ & w/o $f_{\text{3D}}$& w/o $\mathcal{L}_{\text{grad}}$  \\ \hline
    IoU$\uparrow$  & \textbf{0.783} &0.782 &0.718 &0.759\\ 
    Chamfer-L1$\downarrow$  & \textbf{0.058} & 0.060 & 0.082 &0.066\\ 
\bottomrule
\end{tabular}
}
\vspace{-0.1in}
\caption{Ablation study of the decoder design and the gradient loss. We use ShapeNet and ground truth poses in this experiment. The results are averaged across all 13 categories.}
\label{table:ablation-shape-init}
\vspace{-0.1in}
\end{table}

\begingroup
\begin{figure*}

 \def\imw{0.25\textwidth}
 \def\imws{0.12\textwidth}
  \def\imh{0.105\textheight}
 \setlength{\tabcolsep}{1pt}
      \begin{minipage}[t]{0.55\textwidth} 
\begin{tabular}{cccccccccc}

\includegraphics[width=\imws, keepaspectratio]{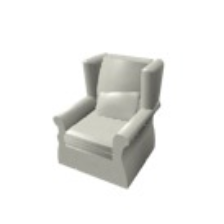} &
\includegraphics[width=\imws, keepaspectratio]{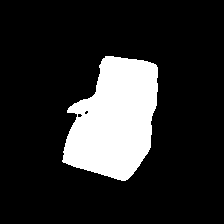} &
\includegraphics[width=\imws, keepaspectratio]{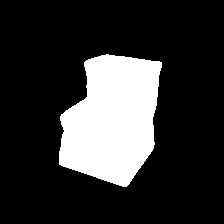} &
\includegraphics[width=\imws, keepaspectratio]{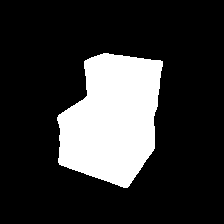} &

\includegraphics[width=\imws, keepaspectratio]{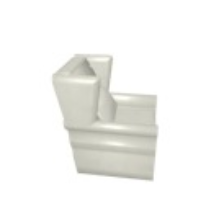} &
\includegraphics[width=\imws, keepaspectratio]{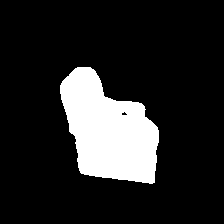} &
\includegraphics[width=\imws, keepaspectratio]{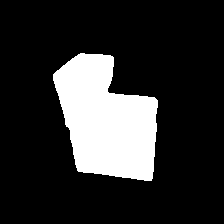} &
\includegraphics[width=\imws, keepaspectratio]{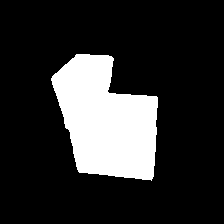}  \\

\includegraphics[width=\imws, keepaspectratio]{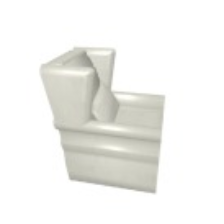} &
\includegraphics[width=\imws, keepaspectratio]{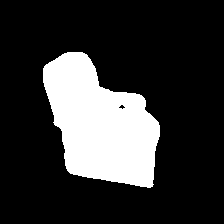} &
\includegraphics[width=\imws, keepaspectratio]{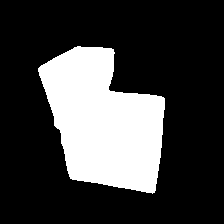} &
\includegraphics[width=\imws, keepaspectratio]{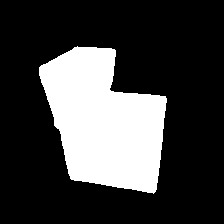} &

\includegraphics[width=\imws, keepaspectratio]{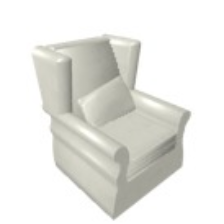} &
\includegraphics[width=\imws, keepaspectratio]{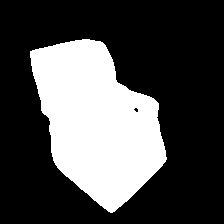} &
\includegraphics[width=\imws, keepaspectratio]{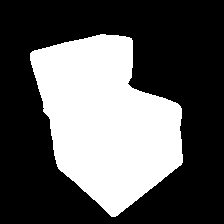} &
\includegraphics[width=\imws, keepaspectratio]{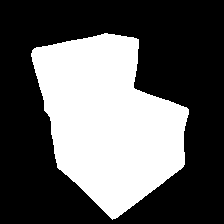} & \\

\includegraphics[width=\imws, keepaspectratio]{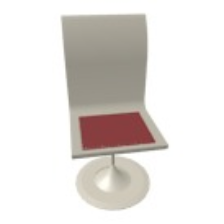} &
\includegraphics[width=\imws, keepaspectratio]{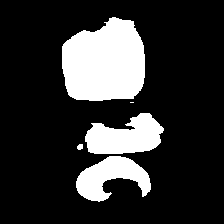} &
\includegraphics[width=\imws, keepaspectratio]{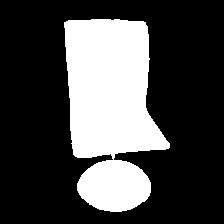} &
\includegraphics[width=\imws, keepaspectratio]{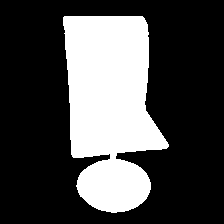} &

\includegraphics[width=\imws, keepaspectratio]{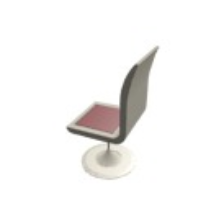} &
\includegraphics[width=\imws, keepaspectratio]{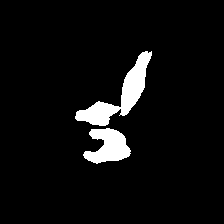} &
\includegraphics[width=\imws, keepaspectratio]{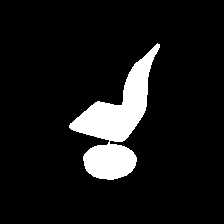} &
\includegraphics[width=\imws, keepaspectratio]{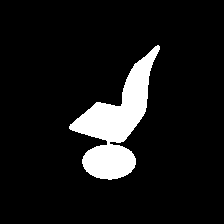}  \\

\includegraphics[width=\imws, keepaspectratio]{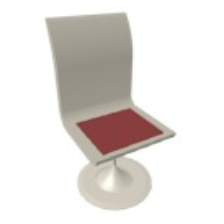} &
\includegraphics[width=\imws, keepaspectratio]{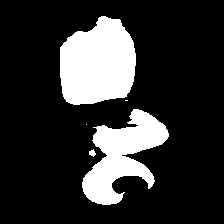} &
\includegraphics[width=\imws, keepaspectratio]{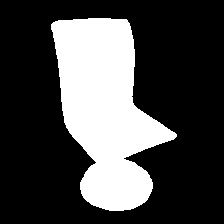} &
\includegraphics[width=\imws, keepaspectratio]{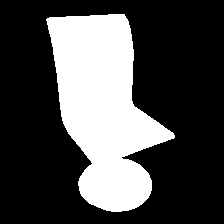} &

\includegraphics[width=\imws, keepaspectratio]{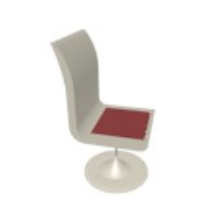} &
\includegraphics[width=\imws, keepaspectratio]{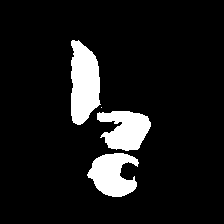} &
\includegraphics[width=\imws, keepaspectratio]{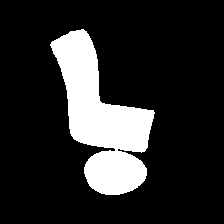} &
\includegraphics[width=\imws, keepaspectratio]{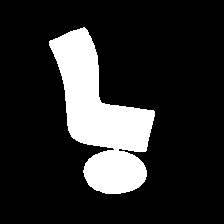} & \\

\includegraphics[width=\imws, keepaspectratio]{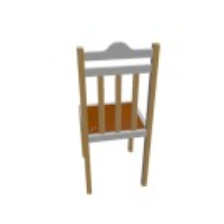} &
\includegraphics[width=\imws, keepaspectratio]{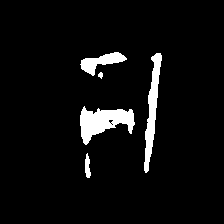} &
\includegraphics[width=\imws, keepaspectratio]{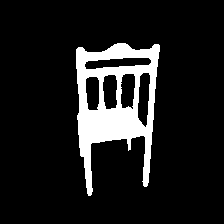} &
\includegraphics[width=\imws, keepaspectratio]{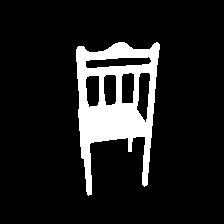} &

\includegraphics[width=\imws, keepaspectratio]{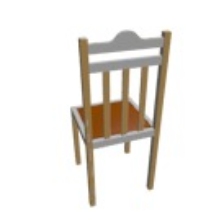} &
\includegraphics[width=\imws, keepaspectratio]{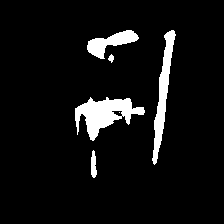} &
\includegraphics[width=\imws, keepaspectratio]{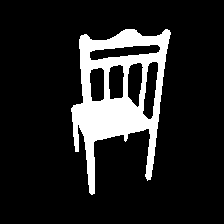} &
\includegraphics[width=\imws, keepaspectratio]{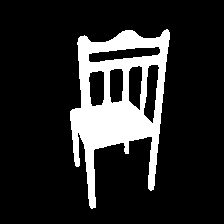}  \\

\includegraphics[width=\imws, keepaspectratio]{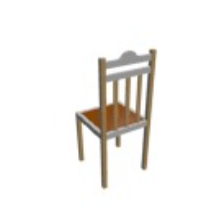} &
\includegraphics[width=\imws, keepaspectratio]{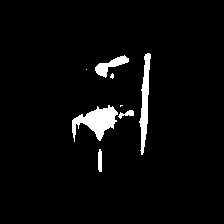} &
\includegraphics[width=\imws, keepaspectratio]{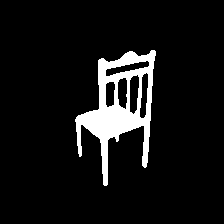} &
\includegraphics[width=\imws, keepaspectratio]{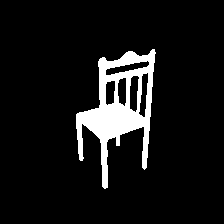} &

\includegraphics[width=\imws, keepaspectratio]{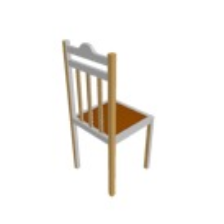} &
\includegraphics[width=\imws, keepaspectratio]{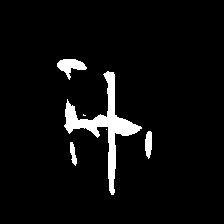} &
\includegraphics[width=\imws, keepaspectratio]{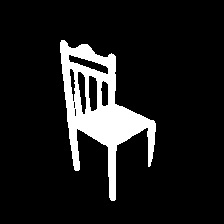} &
\includegraphics[width=\imws, keepaspectratio]{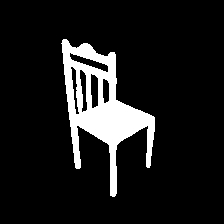} & \\

\includegraphics[width=\imws, keepaspectratio]{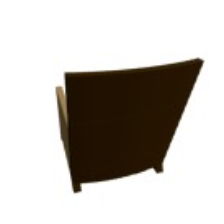} &
\includegraphics[width=\imws, keepaspectratio]{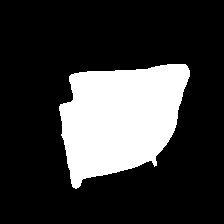} &
\includegraphics[width=\imws, keepaspectratio]{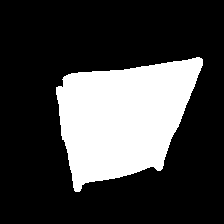} &
\includegraphics[width=\imws, keepaspectratio]{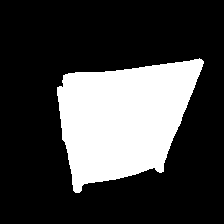} &
\includegraphics[width=\imws, keepaspectratio]{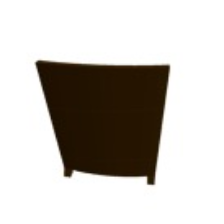} &
\includegraphics[width=\imws, keepaspectratio]{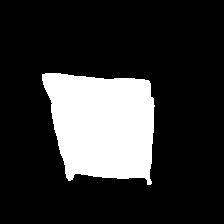} &
\includegraphics[width=\imws, keepaspectratio]{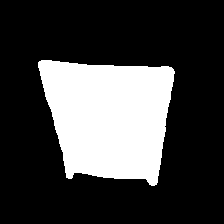} &
\includegraphics[width=\imws, keepaspectratio]{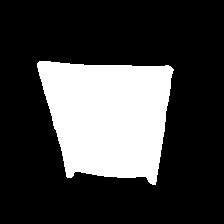} \\
\includegraphics[width=\imws, keepaspectratio]{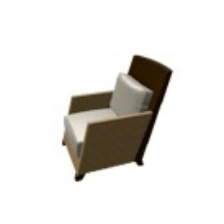} &
\includegraphics[width=\imws, keepaspectratio]{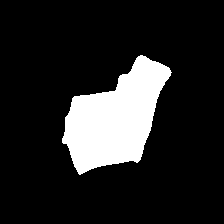} &
\includegraphics[width=\imws, keepaspectratio]{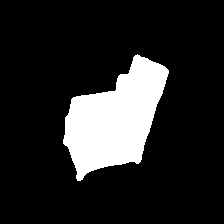} &
\includegraphics[width=\imws, keepaspectratio]{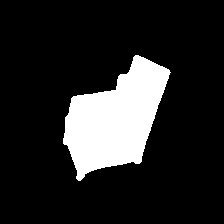} &
\includegraphics[width=\imws, keepaspectratio]{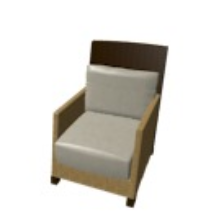} &
\includegraphics[width=\imws, keepaspectratio]{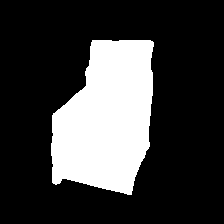} &
\includegraphics[width=\imws, keepaspectratio]{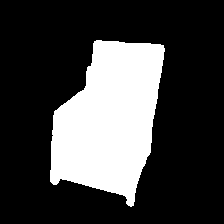} &
\includegraphics[width=\imws, keepaspectratio]{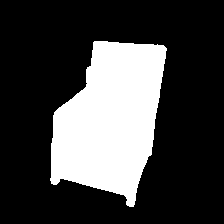} \\

\includegraphics[width=\imws, keepaspectratio]{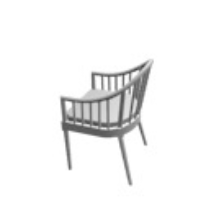} &
\includegraphics[width=\imws, keepaspectratio]{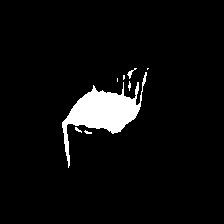} &
\includegraphics[width=\imws, keepaspectratio]{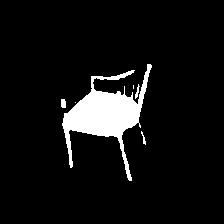} &
\includegraphics[width=\imws, keepaspectratio]{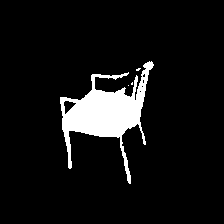} &
\includegraphics[width=\imws, keepaspectratio]{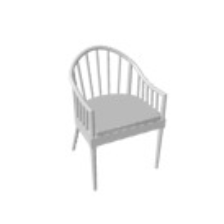} &
\includegraphics[width=\imws, keepaspectratio]{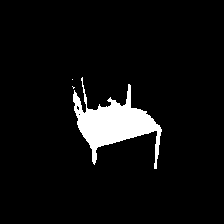} &
\includegraphics[width=\imws, keepaspectratio]{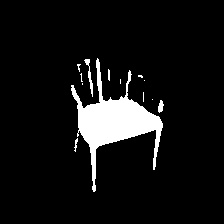} &
\includegraphics[width=\imws, keepaspectratio]{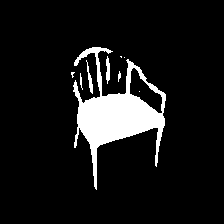} \\
\includegraphics[width=\imws, keepaspectratio]{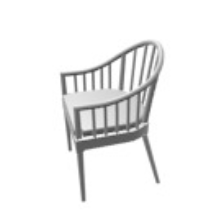} &
\includegraphics[width=\imws, keepaspectratio]{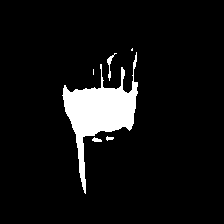} &
\includegraphics[width=\imws, keepaspectratio]{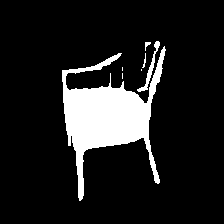} &
\includegraphics[width=\imws, keepaspectratio]{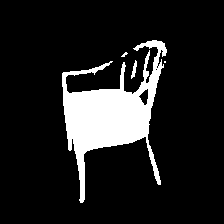} &
\includegraphics[width=\imws, keepaspectratio]{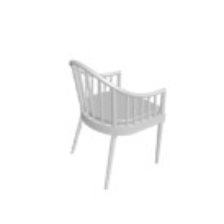} &
\includegraphics[width=\imws, keepaspectratio]{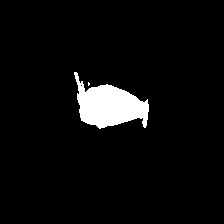} &
\includegraphics[width=\imws, keepaspectratio]{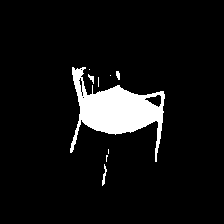} &
\includegraphics[width=\imws, keepaspectratio]{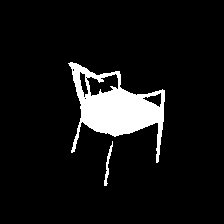} \\

Img & Iter 1 & Iter 2 & Iter 3 & Img & Iter 1 & Iter 2 & Iter 3
\end{tabular}
\end{minipage}
     \begin{minipage}[t]{0.45\textwidth} 
      \setlength{\tabcolsep}{2pt}
\begin{tabular}{cccc}
\includegraphics[height=\imh,keepaspectratio]{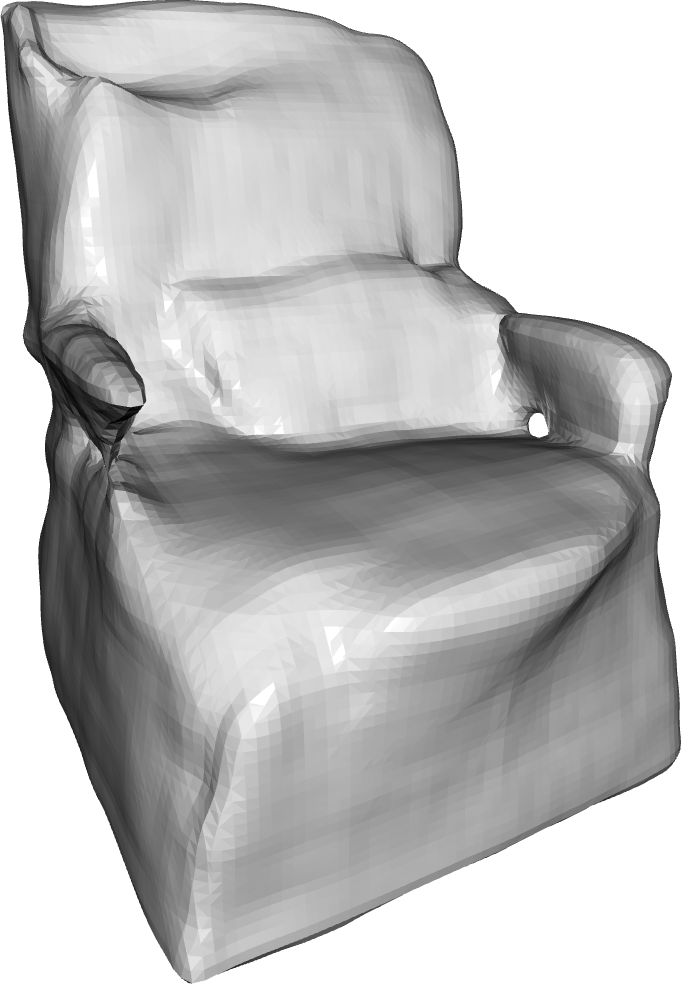} &
\includegraphics[height=\imh,keepaspectratio]{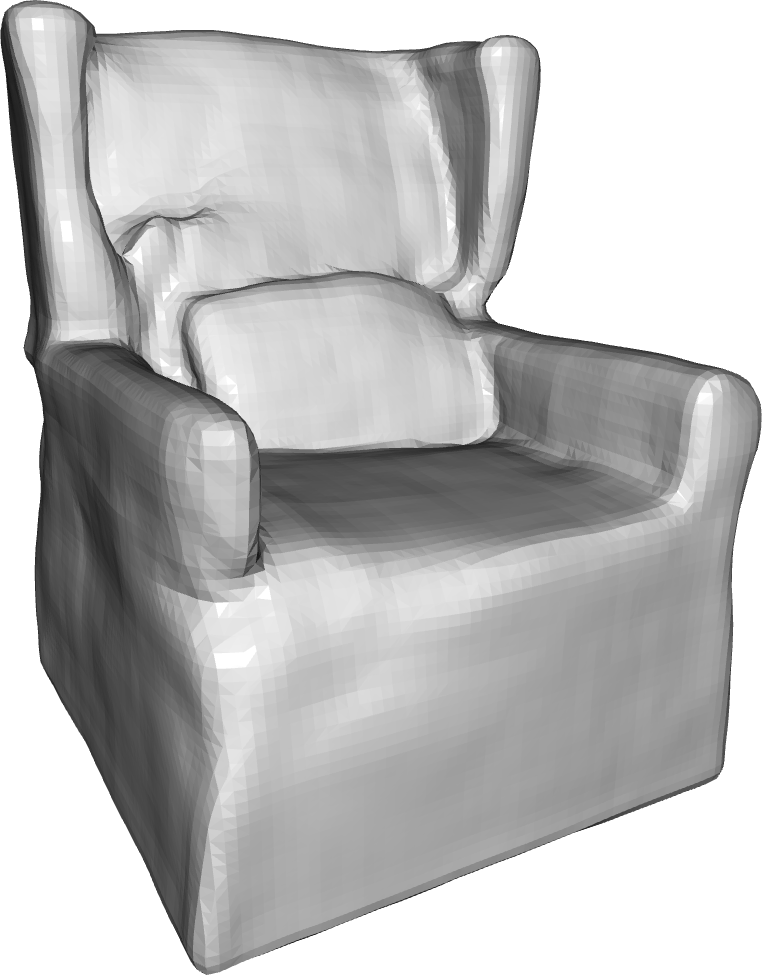} &
\includegraphics[height=\imh,keepaspectratio]{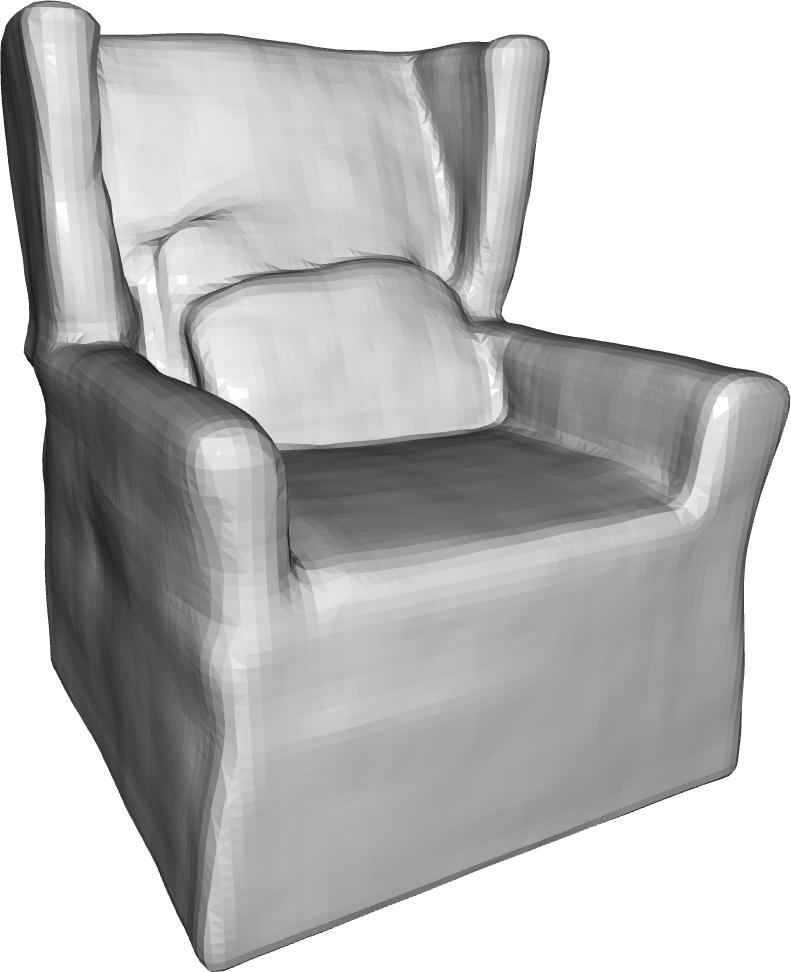} &
\includegraphics[height=\imh,keepaspectratio]{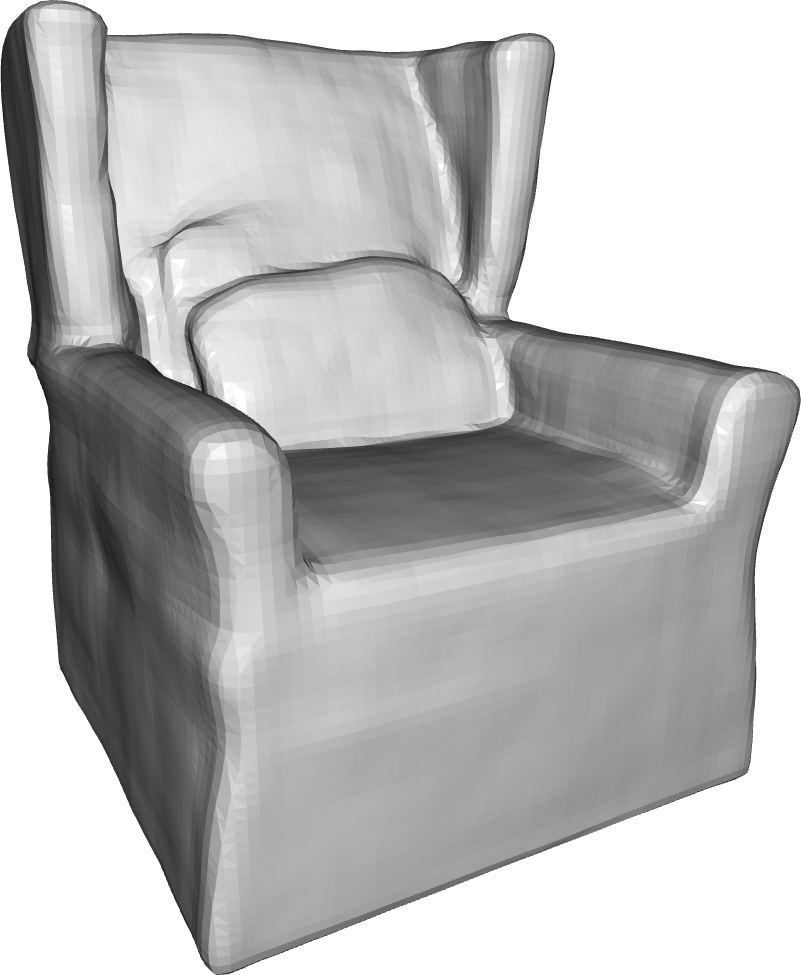} \\

\includegraphics[height=\imh,keepaspectratio]{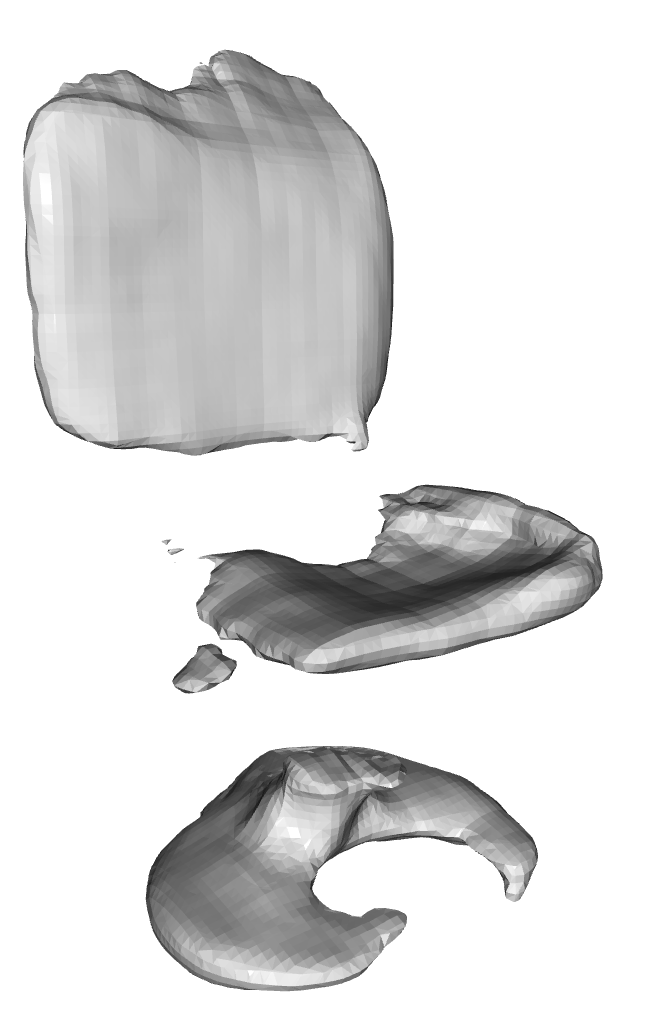} &
\includegraphics[height=\imh,keepaspectratio]{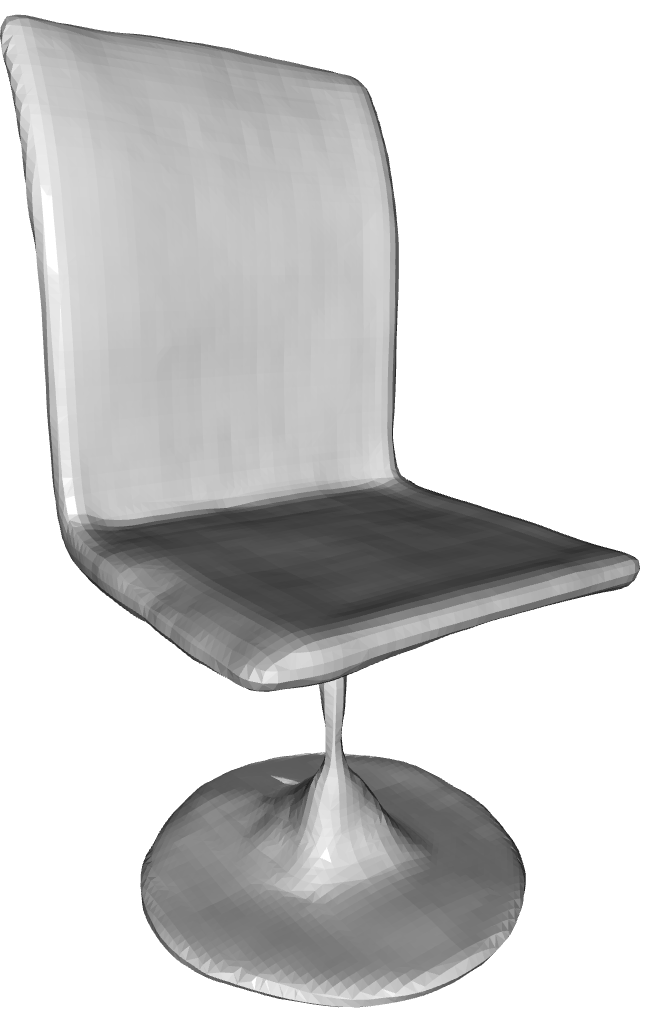} &
\includegraphics[height=\imh,keepaspectratio]{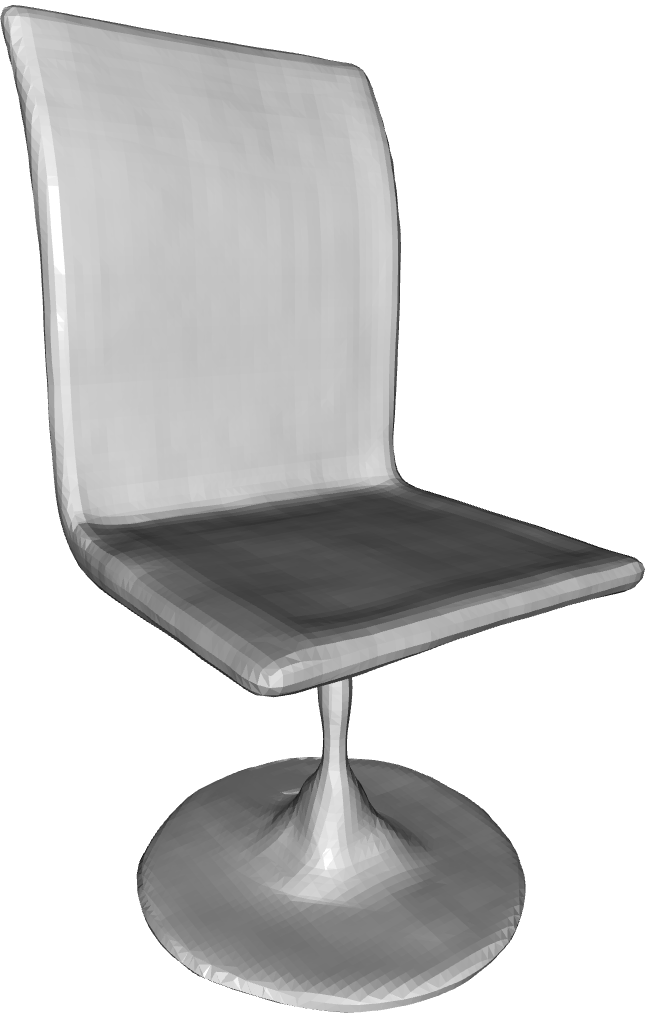} &
\includegraphics[height=\imh,keepaspectratio]{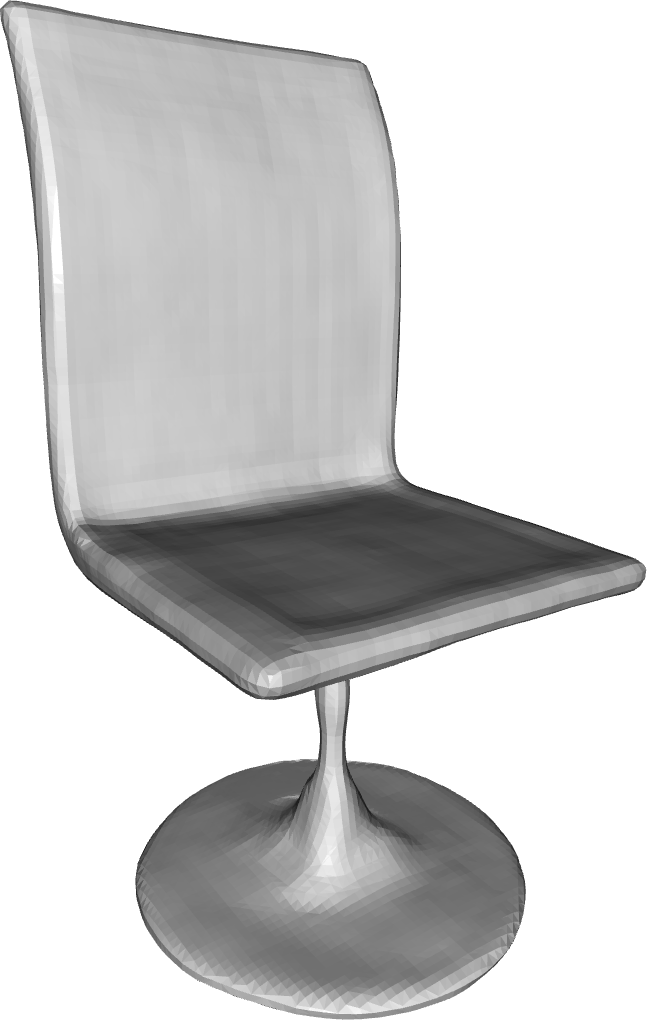} \\

\includegraphics[height=\imh,keepaspectratio]{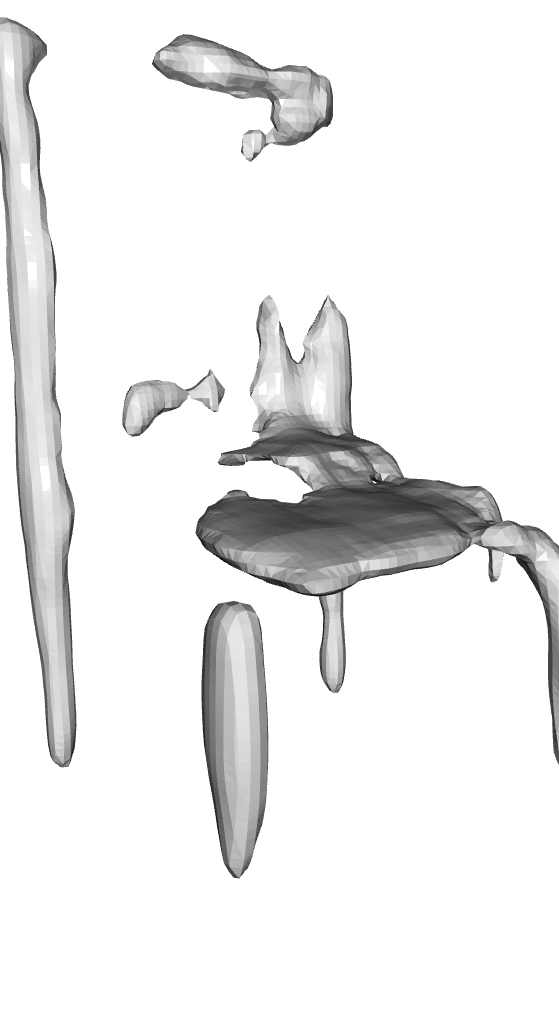} &
\includegraphics[height=\imh,keepaspectratio]{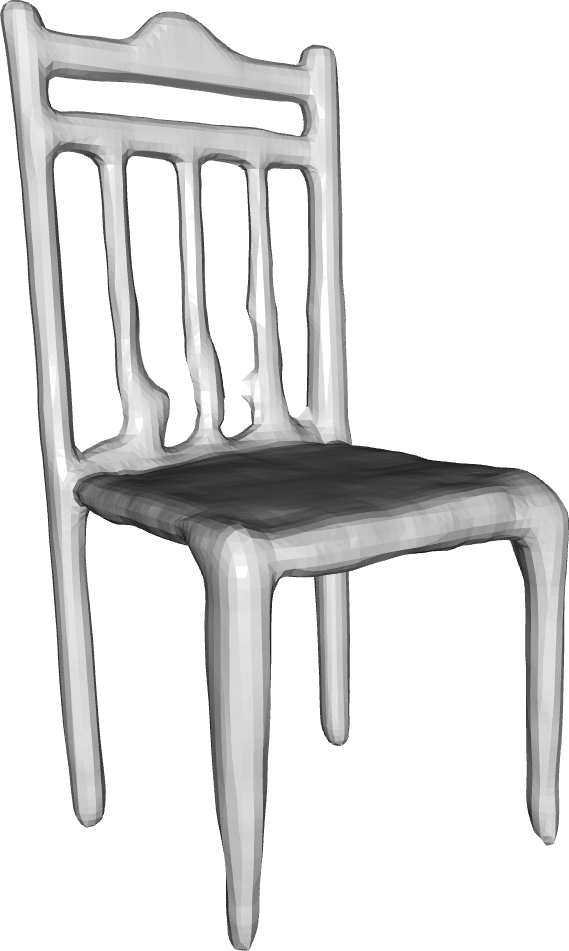} &
\includegraphics[height=\imh,keepaspectratio]{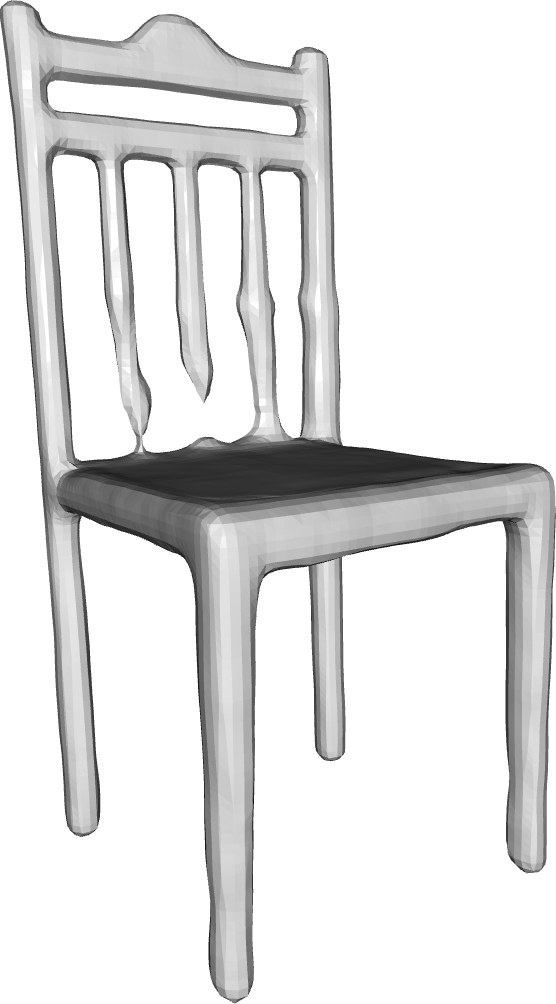} &
\includegraphics[height=\imh,keepaspectratio]{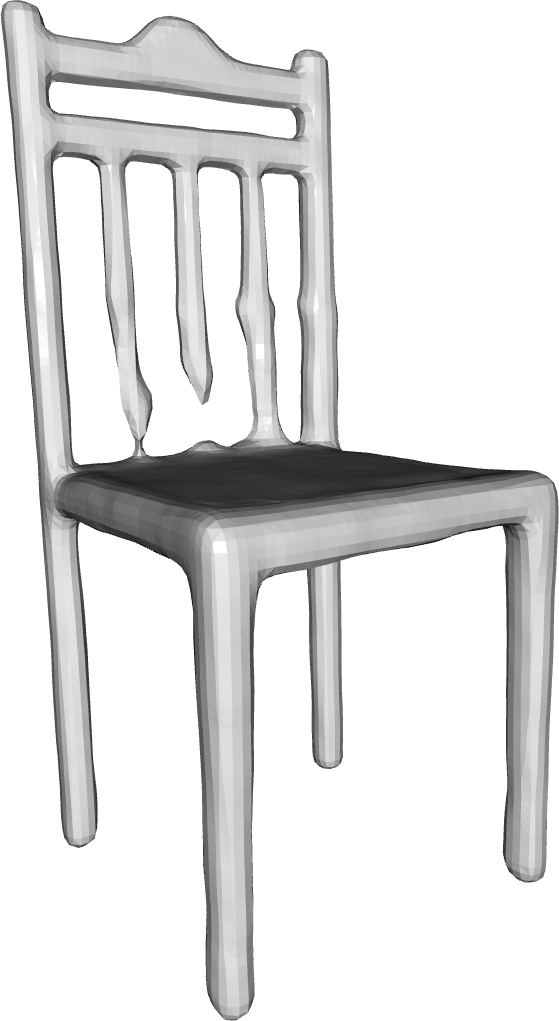} \\

\includegraphics[height=\imh, width=\imw, keepaspectratio]{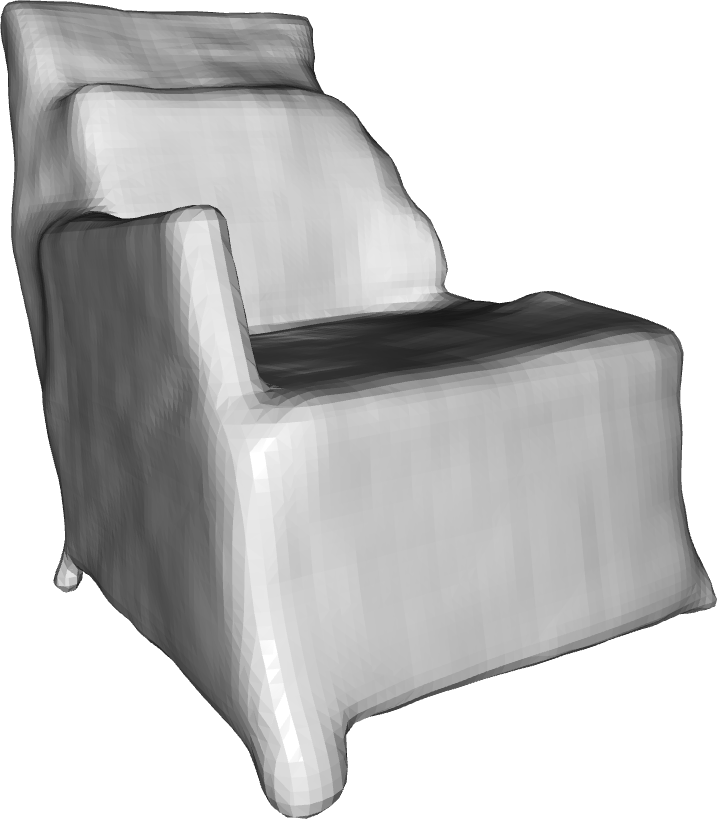}&
\includegraphics[height=\imh, width=\imw, keepaspectratio]{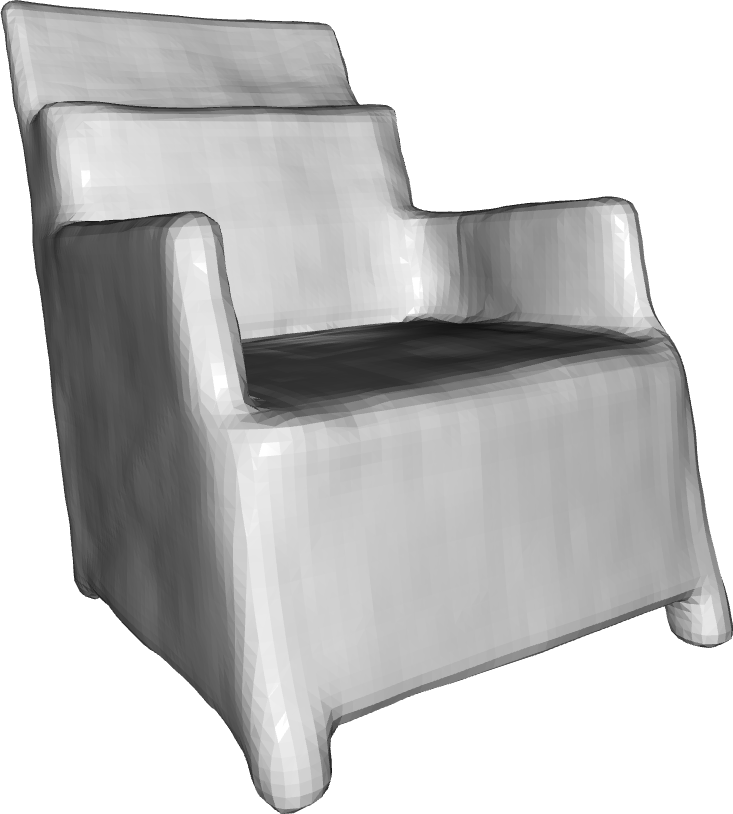}&
\includegraphics[height=\imh, width=\imw, keepaspectratio]{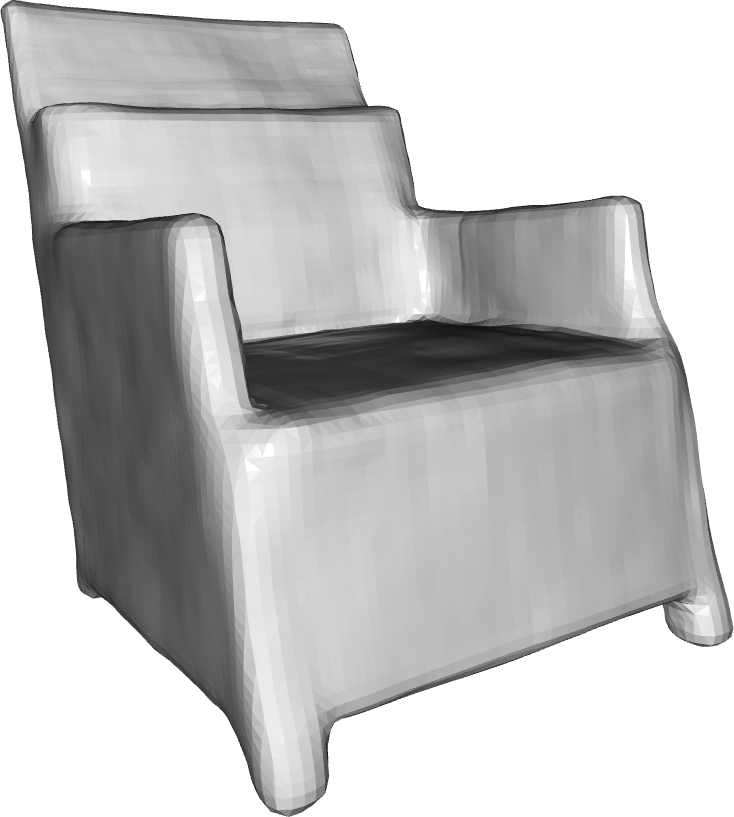}&
\includegraphics[height=\imh, width=\imw, keepaspectratio]{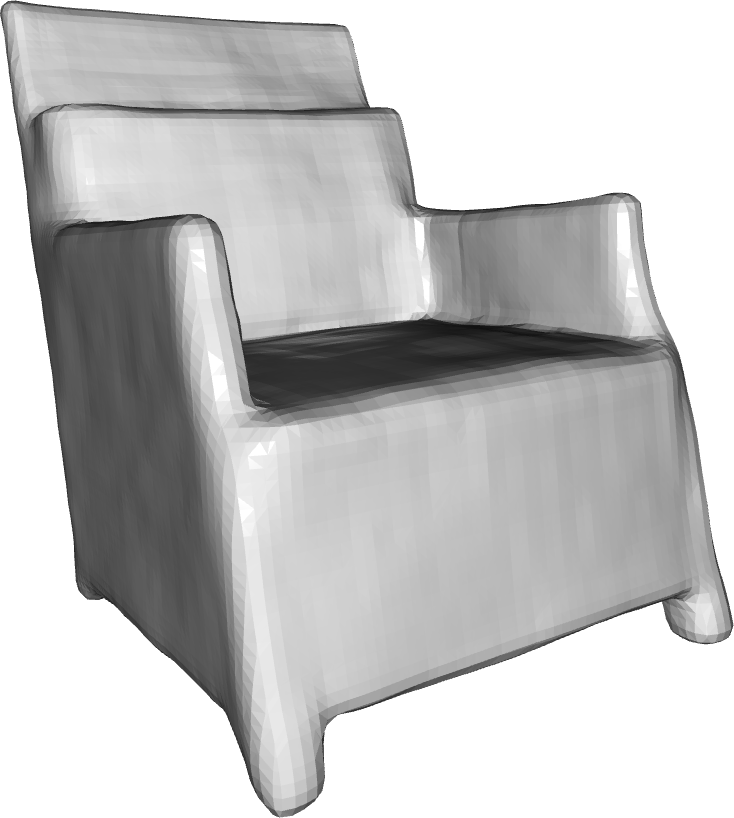}\\

\includegraphics[height=\imh, width=\imw, keepaspectratio]{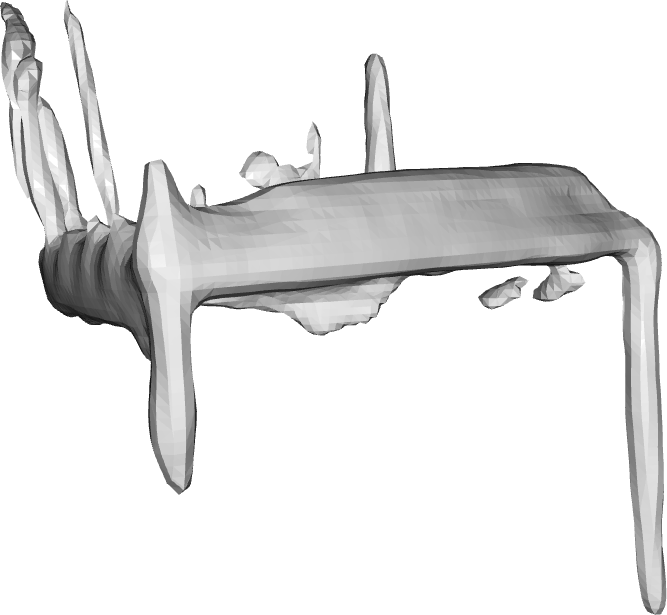}&
\includegraphics[height=\imh, width=\imw, keepaspectratio]{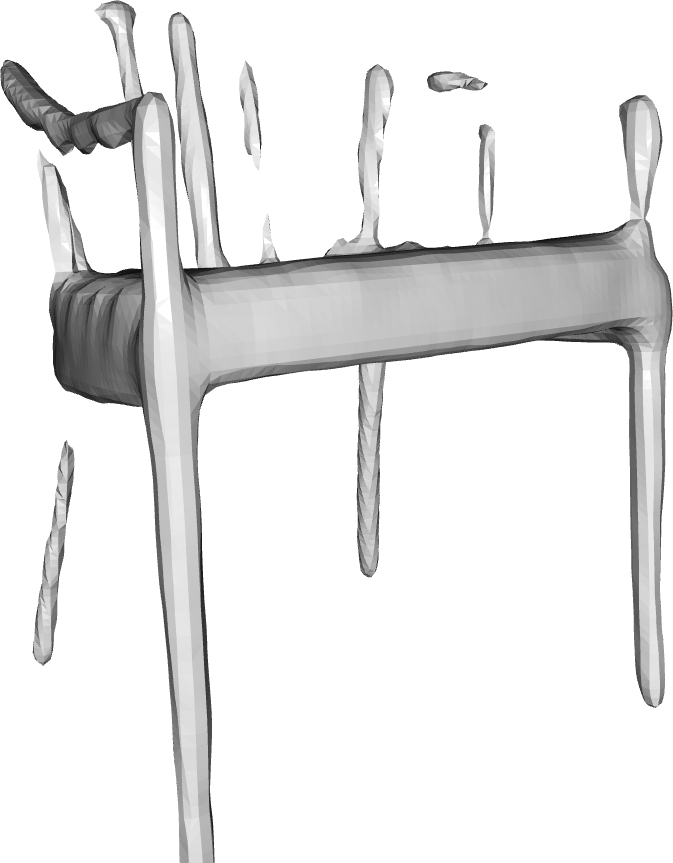}&
\includegraphics[height=\imh, width=\imw, keepaspectratio]{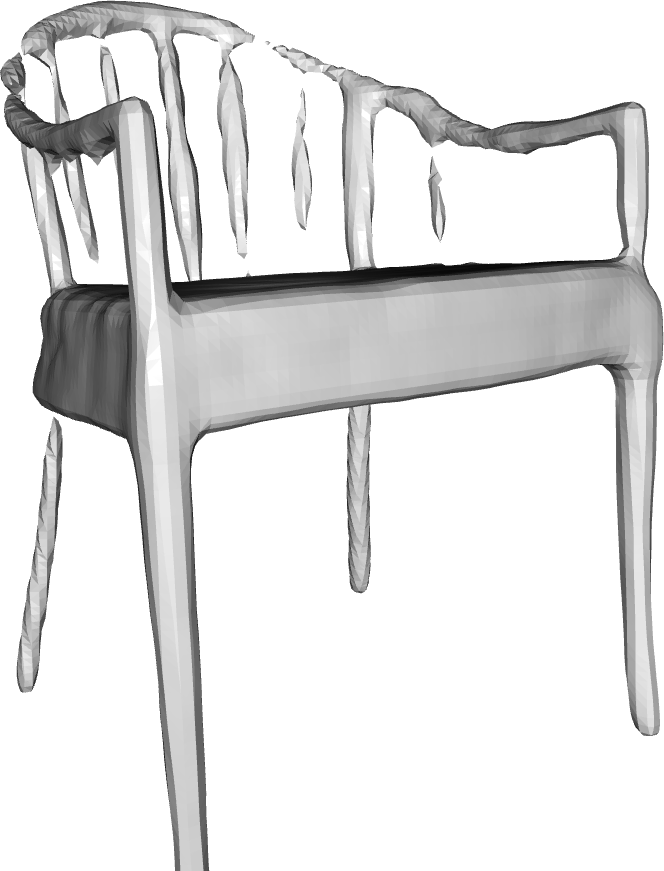}&
\includegraphics[height=\imh, width=\imw, keepaspectratio]{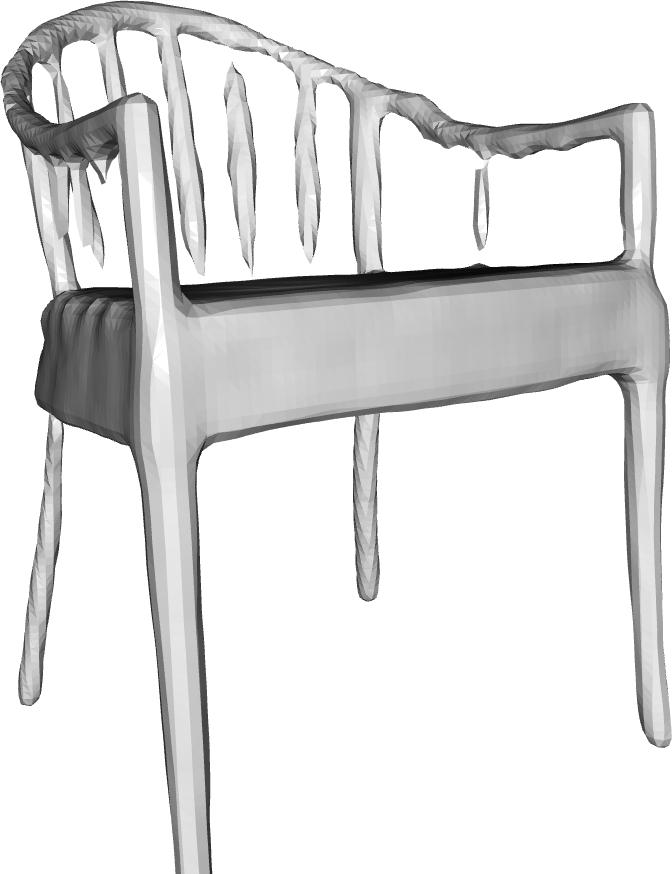}\\

\vspace{-0.5em} Iter 0 & Iter 1 & Iter 2 & Iter 3\\
\end{tabular}
\end{minipage}
\vspace{-0.1in}
\caption{Visualizing the intermediate results of \ouralg. On the left we visualize 4 input views (out of 5 views used) and the back-projected masks in each iteration. On the right we show the progression of 3D reconstructions through alternating shape and pose optimizations.}
\label{fig:shapenet-opt}
\end{figure*}
\endgroup

\noindent\textbf{DISN}~\cite{disn} parametrizes camera poses by orthogonal vectors, and introduces a novel loss for regression-based pose estimation. We implement DISN's camera pose estimation algorithm based on our framework. Note that for baseline comparison, we report results of our updated implementation, which improve from the original results in~\cite{disn}.

\noindent\textbf{Cai~\etal }~\cite{cai2021extreme} Extreme-Rot is a recent deep learning method for estimating pair-wise relative rotations between two images with little overlap. It estimates the rotation by predicting a distribution over discretized Euler angle bins. We implement Extreme-Rot based on our framework and modify it to predict the absolute pose for each image (details are in the supp. material). 

\noindent\textbf{\ouralg-Quat.} 
The above two methods use the continuous rotation matrix and discrete Euler angle representations. In addition, we add a baseline that predicts a quaternion/translation vector. It shares the same backbone as our method. But instead of predicting per-pixel scene coordinates, it averages the feature map of each image to a single vector and regresses the quaternion and translation. 

\subsection{Baselines for 3D Reconstruction}
\label{sec:baseline-shape}
We now describe the 3D reconstruction baselines that we use to evaluate the effectiveness of our approach.

\begin{table*}[!ht]
\centering
\resizebox{0.98\textwidth}{!}{%
\begin{tabular}{r|cccc|cc}
\toprule
\multirow{2}{*}{Category} & \multicolumn{4}{c|}{w/o GT  Pose}&  \multicolumn{2}{c}{w/ GT  Pose} \\
\cline{2-7}
&OccNet~\cite{mescheder2019occupancy} & OccNet$^\dagger$~\cite{apple}&Pix2Vox++ & \ouralg & 3D43D~\cite{apple} & \ouralg  w/ GT Pose \\ \hline
plane    &0.591 / 0.134 / 0.845& 0.600 / 0.096 / 0.853 &0.366/-/-& 0.726 / 0.0541 / 0.905& 0.736 / 0.021 / 0.899 &0.802 / 0.032 / 0.930\\ 
bench    &0.492 / 0.150 / 0.814& 0.547 / 0.176 / 0.834 &0.328/-/-& 0.654 / 0.0530 / 0.904 & 0.663 / 0.027 / 0.881 &0.710 / 0.047 / 0.913\\ 
cabinet    &0.750 / 0.153 / 0.884& 0.770 / 0.125 / 0.893 &0.601/-/- &0.844 / 0.0773 / 0.931&0.831 / 0.073 / 0.925 &0.83 / 0.067 / 0.936\\ 
car    &0.746 / 0.149 / 0.852& 0.759 / 0.109 / 0.861&0.581/-/-&0.768 / 0.0953 / 0.874& 0.797 / 0.090 / 0.873 &0.800 / 0.088 / 0.882\\ 
chair   &0.530 / 0.206 / 0.829 & 0.568 / 0.1870.846 &0.430/-/-& 0.690 / 0.0730 / 0.918&0.716 / 0.063 / 0.911 &0.746 / 0.057 / 0.932\\ 
display   &0.518 / 0.258 / 0.857 & 0.593 / 0.168 / 0.884 &0.443/-/-&0.754 / 0.0769 / 0.935&0.752 / 0.089 / 0.935&0.794 / 0.058 / 0.950\\ 
lamp   &0.400 / 0.368 / 0.751 & 0.415/ 1.083 / 0.764 &0.277/-/- &0.599 / 0.116 / 0.867&0.625 / 0.256 / 0.858&0.682 / 0.069 / 0.893\\ 
speaker   &0.677 / 0.266 / 0.848  & 0.699 / 0.360 / 0.856 &0.588/-/- &0.793 / 0.109 / 0.908&0.807 / 0.143 / 0.912&0.807 / 0.089 / 0.919\\ 
rifle   &0.480 / 0.143 / 0.783  & 0.466 / 0.112 / 0.789 &0.338/-/-&0.705 / 0.0476 / 0.913 &0.745 / 0.012 / 0.903&0.823 / 0.0269 / 0.944\\ 
sofa  &0.693 / 0.181 / 0.867  & 0.731 / 0.171 / 0.886&0.554/-/- &0.804 / 0.0748 / 0.938& 0.809 / 0.054 / 0.927 &0.834 / 0.063 / 0.943\\ 
table  &0.542 / 0.182 / 0.860  &0.569 / 0.588 / 0.873 &0.373/-/- &0.654 / 0.0726 / 0.923&0.689 / 0.058 / 0.921&0.706 / 0.060 / 0.934\\ 
phone  &0.740 / 0.127 / 0.939 & 0.785 / 0.103 / 0.948 &0.589/-/- &0.855 / 0.0434 / 0.978&0.861 / 0.017 / 0.971&0.875 / 0.039/ 0.977\\
boat  &0.547 / 0.201 / 0.797 & 0.592 / 0.163 / 0.818  &0.437/-/-&0.712 / 0.0816 / 0.884&0.708 / 0.053 / 0.868&0.763 / 0.064 / 0.906\\ \hline
Mean  & 0.593 / 0.194 / 0.840  & 0.621 / 0.265 / 0.854 & 0.455/-/-& \textbf{0.735} / \textbf{0.075} / \textbf{0.914} &0.749 / 0.073 / 0.906&\textbf{0.783} / \textbf{0.058} / \textbf{0.928}\\ 

\bottomrule
\end{tabular}
}
\caption{Quantitative results of few-view 3D reconstruction on the ShapeNet dataset. The numbers in each cell is (IoU / Chamfer-L1 / F-score). OccNet~\cite{mescheder2019occupancy} uses a single view. The rest of the methods use 5 views. The last two columns show methods that use GT camera poses. We do not factor out similarity because we obtained results for OccNet/OccNet$^\dagger$/3D43D directly from original papers. Chamfer-L1 is multiplied by 10~\cite{mescheder2019occupancy}. }
\label{table:shapenet-shape-init}
\end{table*}

\noindent\textbf{OccNet}~\cite{mescheder2019occupancy} is a top performing method for single-view 3D reconstruction . We follow the practice of 3D43D~\cite{apple} which provides a multi-view augmented version of OccNet denoted as \textbf{OccNet$^\dagger$}~\cite{apple,mescheder2019occupancy}. For Tab.~\ref{table:shapenet-shape-init}, we use the evaluation results provided in 3D43D~\cite{apple}.

\noindent\textbf{Pix2Vox++}~\cite{xie2020pix2vox++}
is a recent work that provides an improved framework to 3D-R2N2~\cite{choy20163d} with multi-scale context-aware fusion. We train their model on ShapeNet using our settings (5 views). We evaluate their prediction against continuous mesh instead of the discretized version~\cite{mescheder2019occupancy}.

\noindent\textbf{3D43D}~\cite{apple}
is a recent work that uses pixel-aligned feature representations and multi-view images with ground-truth camera poses for object 3D reconstruction. 

\noindent\textbf{IDR}~\cite{idr} is an optimization based algorithm that does not learn a prior from training data. IDR achieves good performance when reconstructing objects with tens of images paired with ground truth object masks. We run IDR~\cite{idr} on each test inputs for 1000 epochs on ShapeNet  for best results. 

\noindent\textbf{\ouralg w/ GT Pose} is our standalone 3D reconstruction module trained with ground truth camera poses. This setting is also used in 3D43D~\cite{apple}, but we differ from 3D43D in network architecture. Although DISN~\cite{disn} also shown in qualitative results for multi-view reconstruction in their paper, we do not find their official implementation for multi-view reconstruction. Instead, we compare a variant of our method w/o $f_{\text{image}}$ which is similar to DISN, in Tab.~\ref{table:ablation-shape-init}.

\noindent\textbf{\ouralg + Noise@L\{1,2,3\}.} For this baseline, we train our standalone 3D reconstruction module with noisy input poses. In order to do this, we add Gaussian noise to the camera poses at 3 different levels of standard deviation ($\sigma\in \big\{0.75\mathrm{e}{-2}, 1.5\mathrm{e}{-2}, 2.25\mathrm{e}{-2}\big\}$)~\cite{lin2021barf}. 

\noindent\textbf{\ouralg w/o Joint} is our proposed approach without performing iterative refinement during inference. We use this baseline to demonstrate the importance of pose optimization for improving robustness to noisy poses.

\subsection{Evaluation Metrics}\label{sec:metric}
\noindent\textbf{Pose Initialization.} We evaluate the camera pose estimation accuracy using three metrics. The first metric is \textit{Pixel-Error}, which is calculated by first projecting the object's surface point into each view using predicted pose and GT pose, and then calculating the corresponding distance in the pixel space. The other two metrics are \textit{Rotation Error} and \textit{Translation error}.  \textit{Pixel-Error} is a more reasonable metric for evaluating poses for multi-view 3D reconstruction, as it reflects both the rotation and translation errors~\cite{disn}.

\noindent\textbf{3D Reconstruction} We measure the distance between a predicted 3D mesh $\hat{S}$ and a ground truth 3D mesh $S$ using common metrics~\cite{mescheder2019occupancy,apple,disn} including \textit{IoU}, \textit{Chamfer-L1 distance}, and \textit{normal consistency}. To eliminate the influence of predicted poses on the final mesh reconstruction, for each method we factor out the similarity transformation between the reconstructed mesh and the underlying ground-truth mesh for evaluation in Tab.~\ref{table:shapenet-shape-refine}(details in the supp). However, to provide a direct comparisons with previous approaches, we do not perform this alignment for results reported in Tab.~\ref{table:shapenet-shape-init}).

\subsection{Results Analysis}\label{sec:analysis}

\noindent\textbf{Pose initialization} In Tab.~\ref{table:ablation-pose-init}, we report results for the different pose initialization baselines on the ShapeNet. The per-category results on ShapeNet can be found in the supp. We make two key observations from these results. First, our approach is the top-performing approach on both datasets. In particular, our method has only a $1.40$ pixel error on ShapeNet. Given that all the baselines for pose estimation share the same backbone, these results demonstrate the benefits of our scene-coordinate representation of camera poses. Second, we also observe from Tab.~\ref{table:ablation-pose-init} that removing the cross-view attention module reduces the accuracy of the pose estimates across the board. This illustrates the importance of aggregating information across multiple views.
\begin{figure}
\centering
\includegraphics[width=0.32\linewidth]{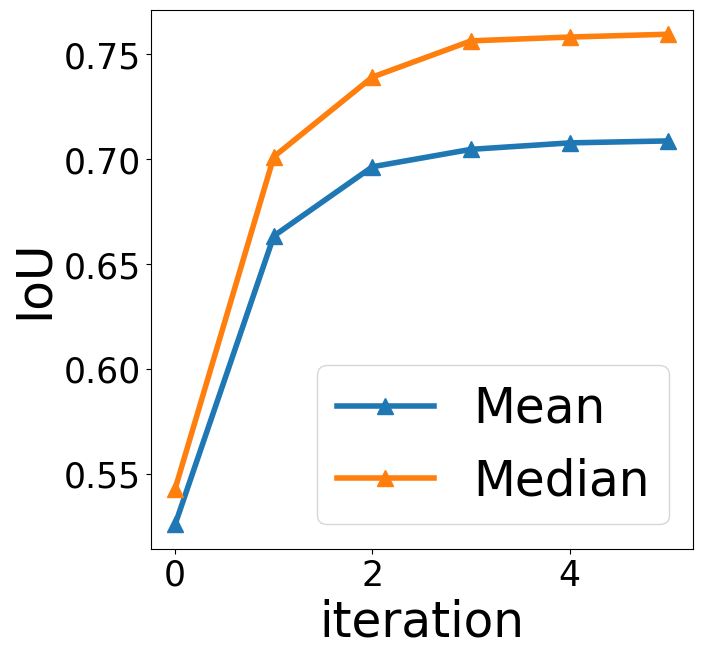} 
\includegraphics[width=0.32\linewidth]{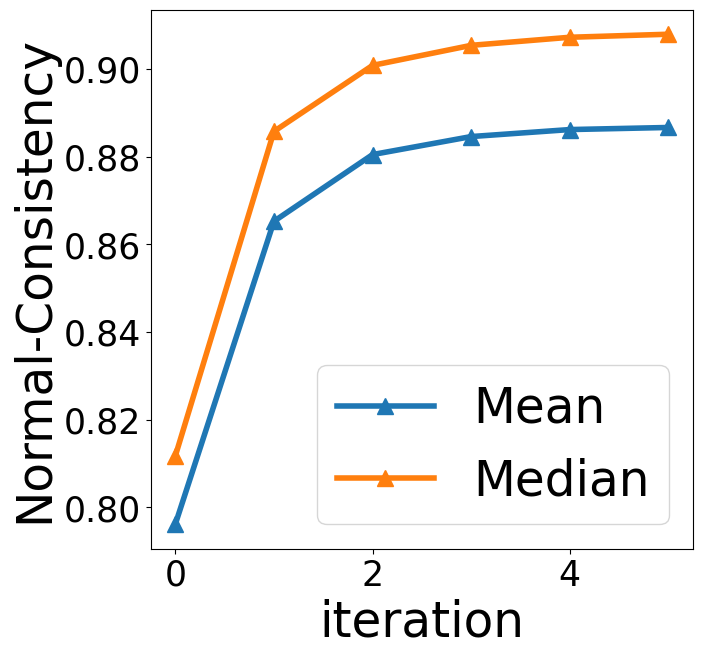} 
\includegraphics[width=0.32\linewidth]{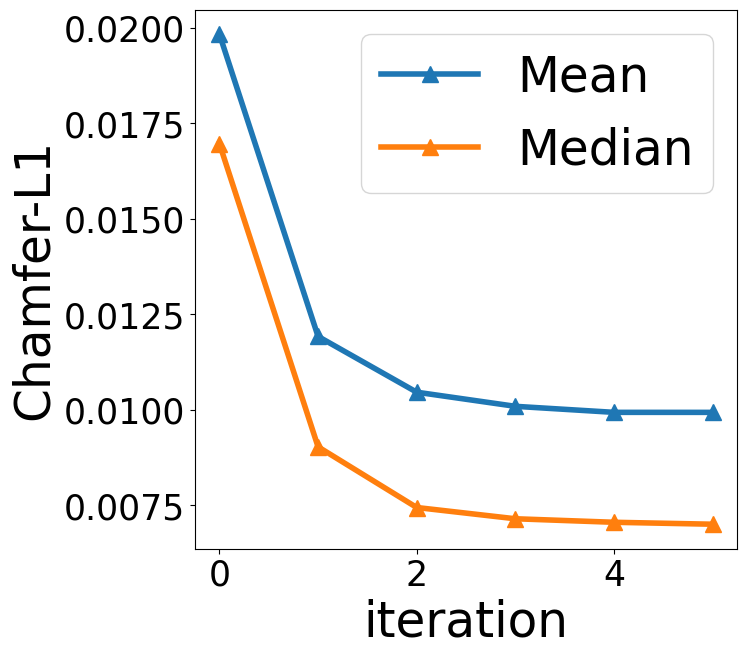} 
\vspace{-0.1in}
\caption{Per iteration results of the refinement approach on ShapeNet under Noise@L3. Our iterative refinement approach has improved 3D reconstruction scores. The refinement process also converges quickly in 3$\sim$ 4 iterations.}
\label{fig:iter}
\end{figure}

\begingroup
\begin{figure*}

 \def\imw{0.25\textwidth}
 \def\imws{0.20\textwidth}
  \def\imh{0.105\textheight}
  \def\imhs{0.08\textheight}
 \setlength{\tabcolsep}{1pt}
      \begin{minipage}[t]{0.55\textwidth} 
\begin{tabular}{ccccc}

\includegraphics[width=\imws, keepaspectratio]{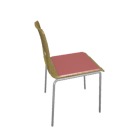} &
\includegraphics[width=\imws, keepaspectratio]{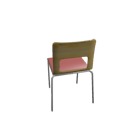} &
\includegraphics[width=\imws, keepaspectratio]{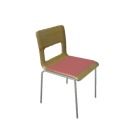} &
\includegraphics[width=\imws, keepaspectratio]{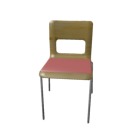} &
\includegraphics[width=\imws, keepaspectratio]{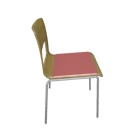}\\

\includegraphics[width=\imws, keepaspectratio]{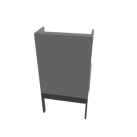} &
\includegraphics[width=\imws, keepaspectratio]{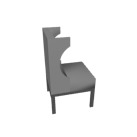} &
\includegraphics[width=\imws, keepaspectratio]{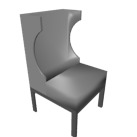} &
\includegraphics[width=\imws, keepaspectratio]{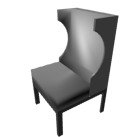} &
\includegraphics[width=\imws, keepaspectratio]{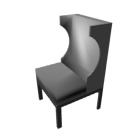} \\

\includegraphics[width=\imws, keepaspectratio]{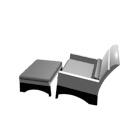} &
\includegraphics[width=\imws, keepaspectratio]{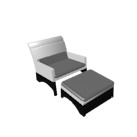} &
\includegraphics[width=\imws, keepaspectratio]{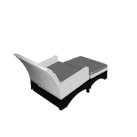} &
\includegraphics[width=\imws, keepaspectratio]{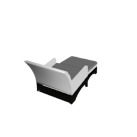} &
\includegraphics[width=\imws, keepaspectratio]{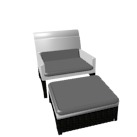} \\

Img0 & Img1 & Img2 & Img3 & Img4
\end{tabular}

\end{minipage}
     \begin{minipage}[t]{0.45\textwidth} 
      \setlength{\tabcolsep}{8pt}
\begin{tabular}{ccc}

\includegraphics[height=\imh,keepaspectratio]{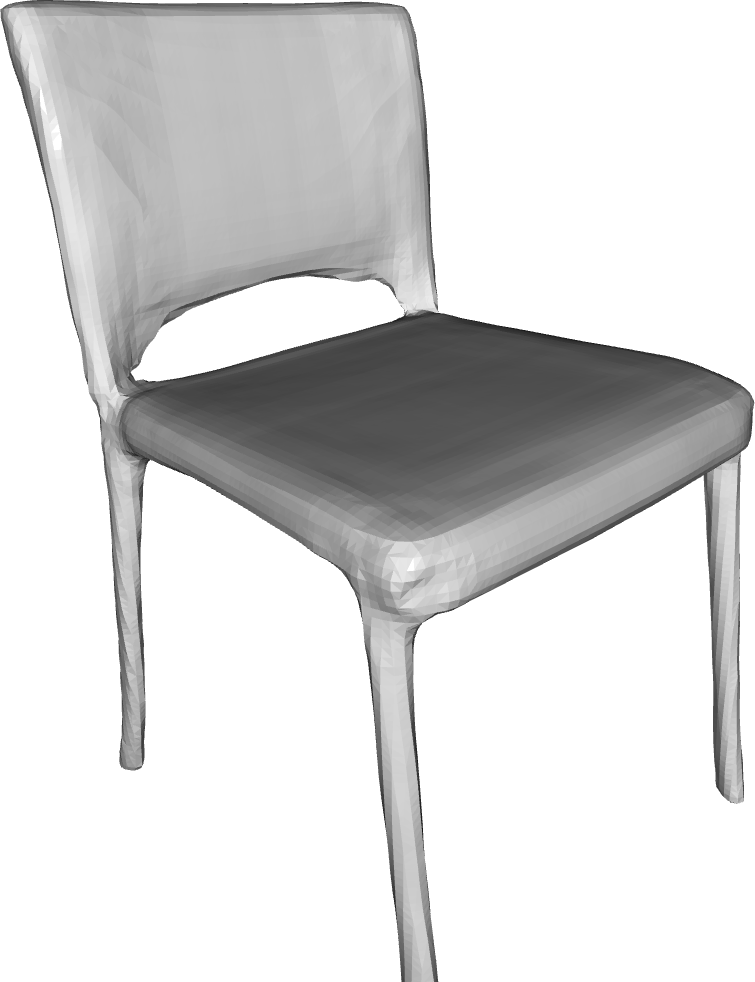} &
\includegraphics[height=\imh,keepaspectratio]{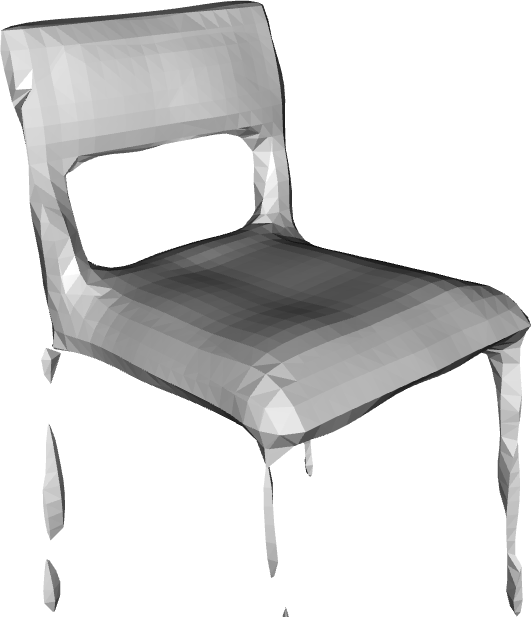} &
\includegraphics[height=\imh,keepaspectratio]{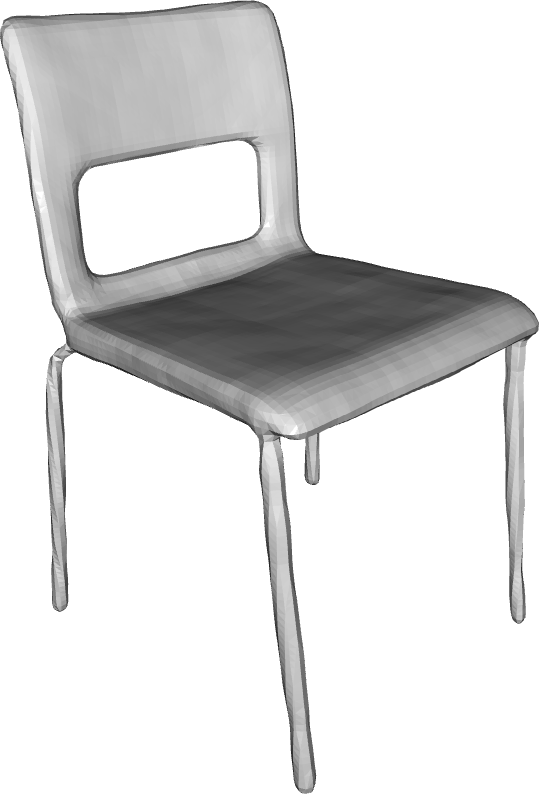} \\

\includegraphics[height=\imh,keepaspectratio]{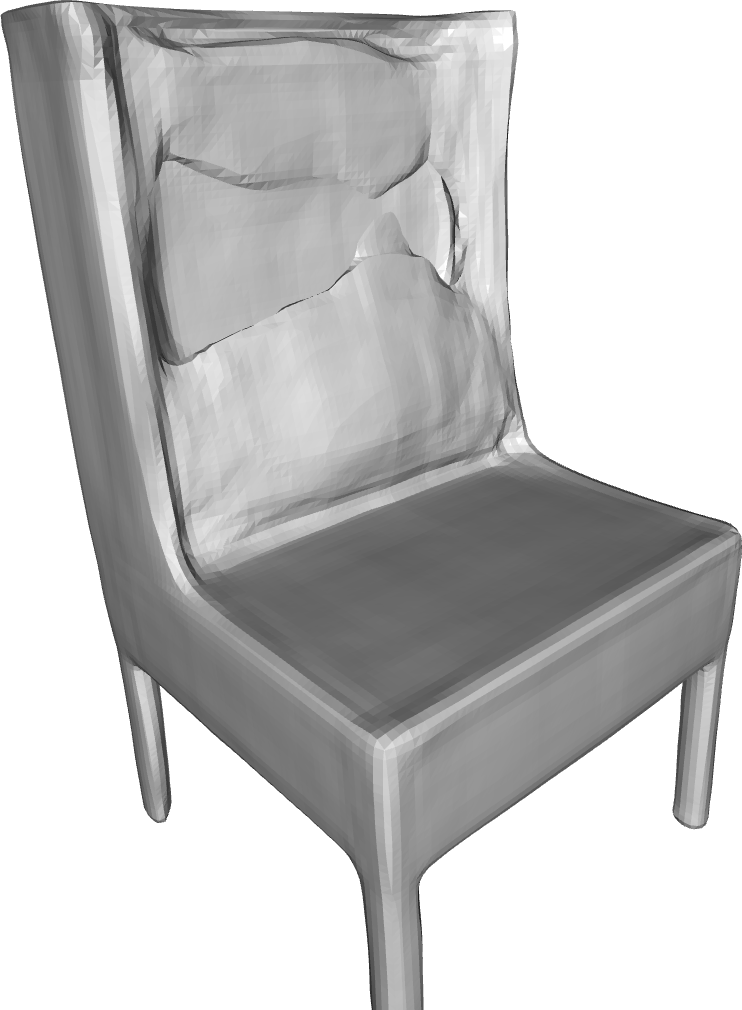} &
\includegraphics[height=\imh,keepaspectratio]{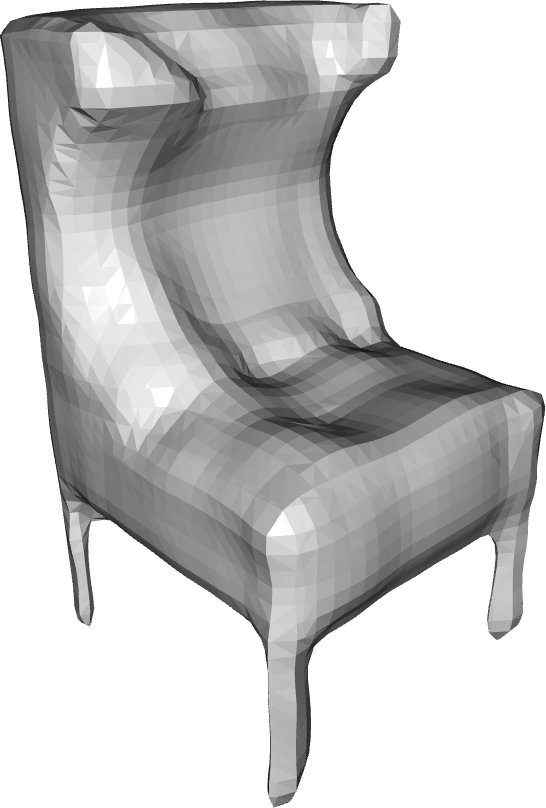} &
\includegraphics[height=\imh,keepaspectratio]{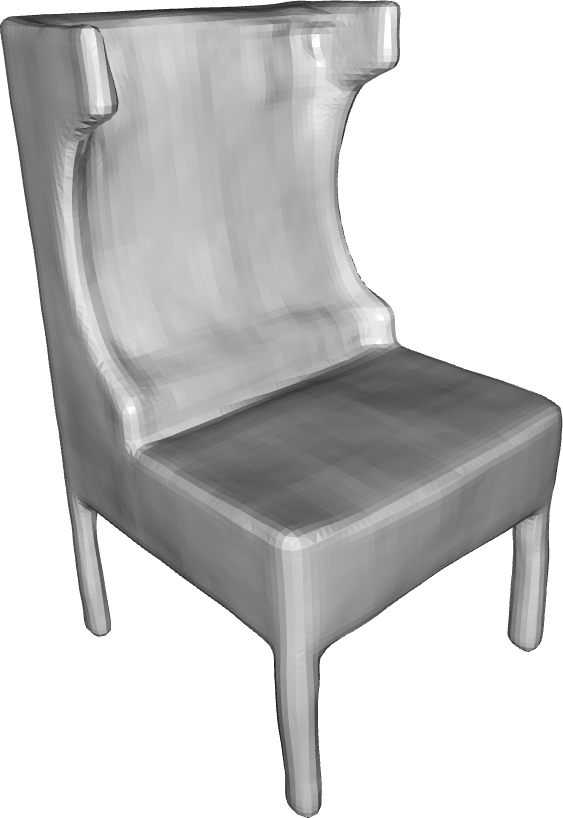} \\

\includegraphics[height=\imhs,keepaspectratio]{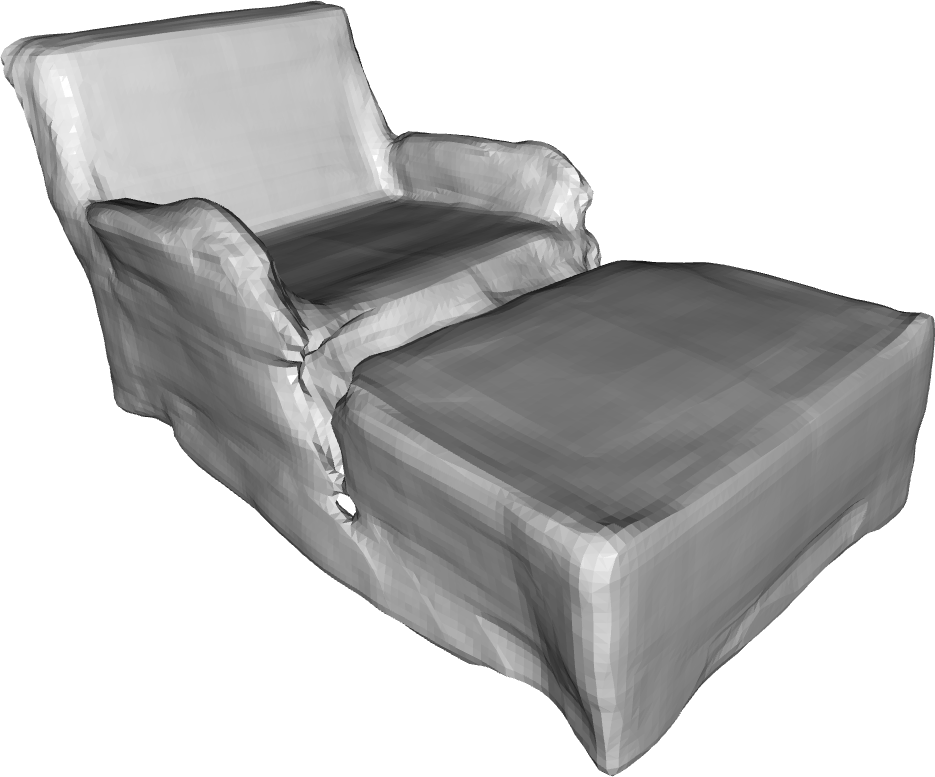} &
\includegraphics[height=\imhs,keepaspectratio]{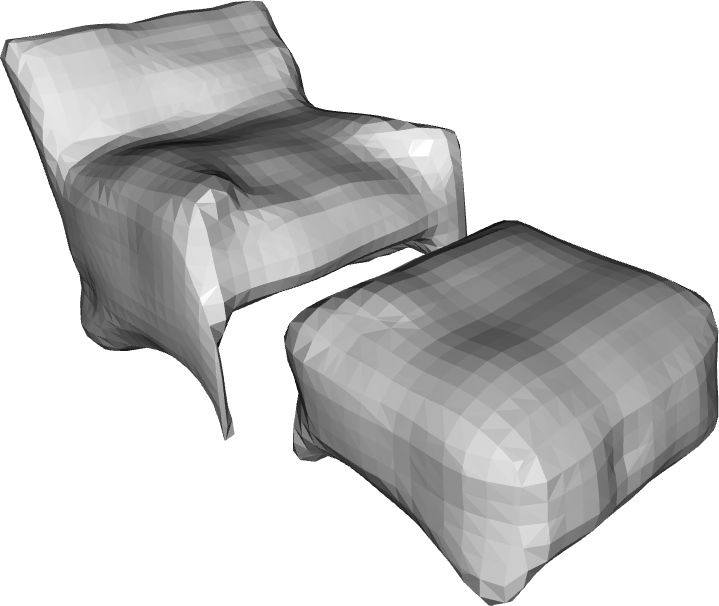} &
\includegraphics[height=\imhs,keepaspectratio]{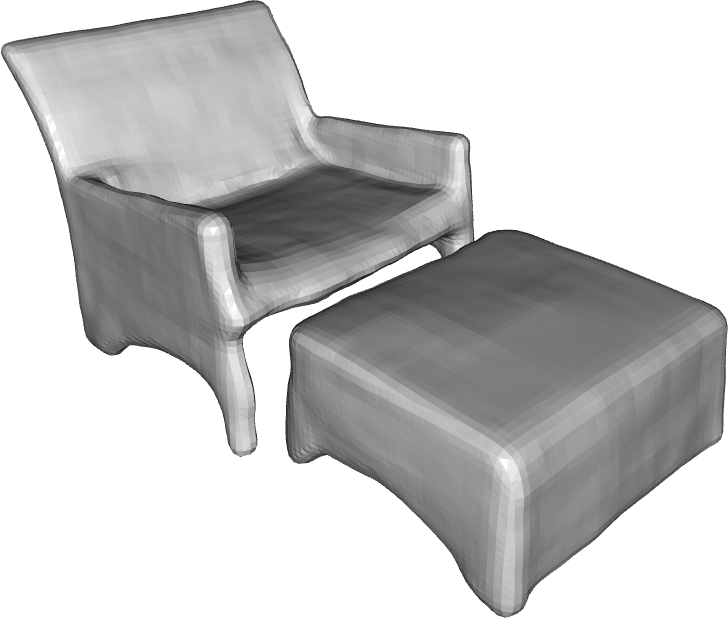} \\

\vspace{-0.5em} OccNet$\dagger$~\cite{apple,mescheder2019occupancy} & IDR~\cite{idr} & FvOR\\
\end{tabular}
\end{minipage}
\vspace{-0.1in}
\caption{Qualitative comparison on ShapeNet. On the left we show the five input images. On the right we show the prediction of OccNet$\dagger$~\cite{apple,mescheder2019occupancy}, IDR~\cite{idr} and FvOR(ours). IDR and FvOR use our predicted camera poses.}
\label{fig:shapenet-compare}
\end{figure*}
\endgroup

\noindent\textbf{Few-view 3D reconstruction w/ GT pose.} To validate our 3D reconstruction module design, we perform multi-view 3D reconstruction experiments on the ShapeNet dataset using ground truth camera poses. The results can be found in Tab.~\ref{table:shapenet-shape-init} (\ouralg $^\dagger$) and Tab.~\ref{table:ablation-shape-init}. Tab.~\ref{table:shapenet-shape-init} shows that our method is the top performer, achieving a $0.783$ IoU,  which is a $4.5\%$ relative improvement compared with the previous state-of-the-art. In addition, the ablation results in Tab.~\ref{table:ablation-shape-init} show that removing the pixel-aligned feature ($f_{\text{image}}$), 3D convolutional feature ($f_{\text{3D}}$) or removing the gradient loss ($\mathcal{L}_{\text{grad}}$) have negative impacts on the reconstruction accuracy, both in terms of IoU and Chamfer-L1.

\begin{table*}
\centering
\resizebox{0.99\textwidth}{!}{%
\begin{tabular}{r|cc|cc cc cc|cc}
\toprule
\multirow{2}{*}{Method}   & \multicolumn{2}{c|}{GT}  & \multicolumn{2}{c}{Noise@L1}  & \multicolumn{2}{c}{Noise@L2} & \multicolumn{2}{c|}{Noise@L3} & \multicolumn{2}{c}{Predict}  \\ 
   & \multicolumn{1}{c}{IoU$\uparrow$} & Chamfer-L1$\downarrow$  & \multicolumn{1}{c}{IoU$\uparrow$} & Chamfer-L1$\downarrow$  & \multicolumn{1}{c}{IoU$\uparrow$} & Chamfer-L1$\downarrow$  & \multicolumn{1}{c}{IoU$\uparrow$} & Chamfer-L1$\downarrow$ & \multicolumn{1}{c}{IoU$\uparrow$} & Chamfer-L1$\downarrow$\\ \hline

OccNet$\dagger$~\cite{mescheder2019occupancy,apple}    & 0.667/0.702 &1.22/0.989& 0.667/0.702 &1.22/0.989 & 0.667/0.702 &1.22/0.989 & 0.667/0.702 &1.22/0.989 & 0.667/0.702 &1.22/0.989 \\

IDR~\cite{idr}    & 0.392/0.358 &4.1/2.4& 0.419/0.444 &4.0/2.2& 0.393/0.364& 4.3/2.6 & 0.415/0.396 &3.9/2.2 & 0.392/0.370& 4.3/2.4 \\
\hline

\ouralg w/ Noisy@L1    & 0.760/0.795 &0.744/0.642& 0.743/0.777 &0.811/0.694& 0.688/0.722& 1.07/0.899 & 0.601/0.631 & 1.64/1.33 & 0.751/0.787& 0.796/0.659 \\

\ouralg w/ Noisy@L2    & 0.725/0.757 &0.899/0.741& 0.718/0.748 &0.935/0.777& \underline{0.703}/\underline{0.733}& \underline{0.998}/\underline{0.851} & 0.676/0.710 & 1.14/0.964 & 0.721/0.752 &0.920/0.756  \\

\ouralg w/ Noisy@L3    & 0.704/0.741 &0.977/0.828& 0.698/0.731 &1.01/0.858& 0.689/0.724& 1.06/0.906 & \underline{0.677}/\underline{0.712} & \underline{1.13}/\underline{0.966}&0.700/0.739&1.02/0.834  \\

\ouralg w/ GT Pose  & \textbf{0.806}/\textbf{0.841} & \textbf{0.605}/\textbf{0.498} &0.667/0.699& 1.17/0.985 &0.531/0.557& 2.05/1.70& 0.441/0.443 & 2.89/2.29 & \textbf{0.785}/\textbf{0.825} & \textbf{0.677}/\textbf{0.543}  \\
\hline

\ouralg w/o Joint &  \underline{0.786}/\underline{0.820}& \underline{0.658}/\underline{0.554}& \underline{0.749}/\underline{0.779}& \underline{0.777}/\underline{0.667}& 0.645/0.677& 1.27/1.09&0.533/0.551& 2.06/1.78& \underline{0.775}/\underline{0.812}& \underline{0.702}/\underline{0.573}   \\

\ouralg   & 0.783/0.818& 0.664/0.561& \textbf{0.779}/\textbf{0.814}& \textbf{0.676}/\textbf{0.571}& \textbf{0.766}/\textbf{0.803}& \textbf{0.735}/\textbf{0.600}&\textbf{0.721}/\textbf{0.768}& \textbf{0.988}/\textbf{0.707}& 0.773/\underline{0.812}& 0.708/0.576  \\

\bottomrule
\end{tabular}
}
\vspace{-0.1in}
\caption{Evaluating the robustness of  few-view 3D reconstruction baselines on ShapeNet (mean/median, top-2 results highlighted). We report the results using ground truth poses, perturbed poses with different perturbation levels, and predicted poses from our pose estimation module. Chamfer-L1 is multiplied by 100. Our approach(\ouralg) can strike a balance between being robust to noisy poses and obtaining high reconstruction accuracy. The details of three noise levels can be found in Section~\ref{sec:analysis}. Note that differently from Table~\ref{table:shapenet-shape-init}, here we pre-align the predicted shape with ground truth shape before evaluation to focus on accessing shape quality. Since OccNet$\dagger$ always predicts a unit scale shape. We've factored out the shape scale when computing the Chamfer-L1 metric of OccNet$\dagger$ for a fair comparison.}  
\label{table:shapenet-shape-refine}
\vspace{-0.1in}
\end{table*}

\noindent\textbf{Robustness of few-view 3D reconstruction under noisy poses.} We experiment two settings of camera poses. In the first setting, a Gaussian noise is applied to the ground truth camera poses, following the practice of BARF~\cite{lin2021barf}. The pose perturbation magnitude is controlled by $\sigma\in \big\{0.75\mathrm{e}{-2}, 1.5\mathrm{e}{-2}, 2.25\mathrm{e}{-2}\big\}$ with three values (called L1, L2, and L3 respectively). The corresponding average pixel errors are $\{2.29, 4.58, 6.88\}$ on ShapeNet. In Tab.~\ref{table:shapenet-shape-refine}, we show results on ShapeNet. The first observation is that \ouralg trained with ground truth poses (\ouralg w/ GT Pose) is highly sensitive to noisy pose initialization at inference time. In particular, the average IoU drops from $0.806$ to $0.667$ with L1 noise, and drops further to $0.441$ with L3 noise. A second observation is that \ouralg trained with noisy poses (\ouralg w/ Noise@L$\{1,2,3\}$) (rows 3\textup{--}5 in Tab.~\ref{table:shapenet-shape-refine}) gains robustness at the trained noise level, as expected. For example, \ouralg w/ Noise@L1 achieves an IoU of $0.743$ at test time with noise level L1, far exceeding the IoU of $0.667$ achieved with \ouralg w/ GT Pose. On the other hand, the robustness of these models (\ouralg w/ Noise@L$\{1,2,3\}$) comes at a cost of decreased performance when accurate camera poses are given, which is expected as they simply fit the network to noisy pose without explicitly modeling(\eg the first column in Tab. \ref{table:shapenet-shape-refine}). 

In contrast, \ouralg with joint shape and pose iterative refinement is a lot more robust to noisy poses, while retaining high performance when the poses become accurate. In Fig.~\ref{fig:iter} we show how reconstruction metrics improve as a function of the number of Levenberg–Marquardt updates in the refinement process. Fig.~\ref{fig:shapenet-opt} shows how the rendered masks and geometry improve at each iteration.

In the second setting, we use predicted poses produced by our pose initialization method during inference. This is shown in the last column of Tab. \ref{table:shapenet-shape-refine}. We observe that our iterative refinement approach (\ouralg) does not provide further gains(i.e. \ouralg w/o joint ) in this case(last 2 rows  of Tab.~\ref{table:shapenet-shape-refine}), which is also expected because on ShapeNet dataset the predicted pose are already fairly close to G.T.~\ref{table:ablation-pose-init}, and our pose update module are designed to address considerable pose error. Qualitative comparison between \ouralg and existing methods on the ShapeNet dataset can be found in Fig.~\ref{fig:shapenet-compare}. \ouralg outperforms existing methods significantly.

\begin{table}
\centering
\resizebox{0.9\linewidth}{!}{%
\begin{tabular}{cccc}
\toprule
     Metric & IoU$\uparrow$ & Chamfer-L1$\downarrow$ & Inference Speed(s)$\downarrow$ \\ \hline
     IDR\cite{idr} & 0.392 & 4.33& $1.0 \times 10^3$ \\ 
     FvOR & \textbf{0.773} & \textbf{0.708}& \textbf{9.8}\\ 
\bottomrule
\end{tabular}
}
\vspace{-0.1in}
\caption{Inference speed for ShapeNet dataset. As an optimization based approach, our method is significant faster than IDR~\cite{idr}.
}
\vspace{-0.1in}
\label{table:speed}
\vspace{-0.2in}
\end{table}

\noindent\textbf{Computational speed.} We found our approach typically converges after 3 shape and pose updates, while IDR~\cite{idr}  requires thousands of updates. The inference speed can be found in the Tab. \ref{table:speed}.

\section{Conclusions and Limitations}

\noindent\textbf{Conclusions.} This paper studied the problem of reconstructing a 3D object from a few observations. We proposed a joint pose and shape refinement approach that strikes a balance between being robust to noisy camera poses and producing accurate 3D reconstructions. 

\noindent\textbf{Limitations.} A limitation of our approach is that separate training of shape and pose module may result in sub-optimal performance. Another limitation is  the pose optimization module requires a reasonable initial shape prediction. We plan to address these limitations in future work.

\FloatBarrier
\clearpage
{\small
\bibliographystyle{ieee_fullname}
\bibliography{refs}
}


\end{document}